\documentclass[3p, review]{elsarticle}

\usepackage{graphicx}
\usepackage[linesnumbered,ruled,vlined]{algorithm2e}
\usepackage{makecell}
\usepackage{tabularx}
\usepackage{hhline}
\usepackage{multirow}
\usepackage{adjustbox}
\usepackage{rotating}
\usepackage{longtable}
\usepackage{booktabs}
\usepackage{xcolor, soul}
\sethlcolor{green}
\usepackage{pdflscape}
\usepackage{colortbl}
\usepackage{subcaption}
\usepackage{amsmath}
\usepackage{tikz}
\usepackage{pifont}
\usepackage{doi}
\usepackage{enumitem}
\usepackage{longtable}
\usepackage{lipsum}
\usepackage{longtable}

\usepackage[numbers]{natbib}
\usepackage{lineno,hyperref}
\modulolinenumbers[5]

\usepackage{tikz}

\journal{Computers and Industrial Engineering}

\biboptions{authoryear}

\begin{document}

\begin{frontmatter}

\title{A survey on pioneering metaheuristic algorithms between 2019 and 2024}

\author[aff1]{Tansel Dokeroglu}
\ead{tansel.dokeroglu@tedu.edu.tr}

\author[aff1]{Deniz Canturk}
\ead{deniz.canturk@tedu.edu.tr}

\author[aff2]{Tayfun Kucukyilmaz}
\ead{kucukyilmaz@rsm.nl}

\address[aff1]{TED University, Software Engineering Department, Ankara, Türkiye}
\address[aff2]{Department of Technology and Operations Management, Erasmus University, The Netherlands}



\begin{abstract}
This review examines over 150 new metaheuristics of the last six years (between 2019-2024), underscoring their profound influence and performance. Over the past three decades, more than 500 new metaheuristic algorithms have been proposed, with no slowdown in sight—an overwhelming abundance that complicates the process of selecting and assessing the most effective solutions for complex optimization challenges. Our evaluation centers on pivotal criteria, including annual citation metrics, the breadth of addressed problem types, source code availability, user-friendly parameter configurations, innovative mechanisms and operators, and approaches designed to mitigate traditional metaheuristic issues such as stagnation and premature convergence. We further explore recent high-impact applications of the past six years' most influential 23 metaheuristic algorithms, shedding light on their advantages and limitations, while identifying challenges and potential avenues for future research.
\end{abstract}


\end{frontmatter}


\clearpage
\section{Introduction}\label{intro}

Over the past decade, a remarkable surge of metaheuristic algorithms has redefined the field, making it a challenge to distinguish the most impactful ones \cite{hussain2019metaheuristic,dokeroglu2019survey,agrawal2021metaheuristic}. With innovation accelerating, selecting the most effective algorithms has become increasingly demanding for researchers and practitioners alike. Recognizing this, we conducted an in-depth review of metaheuristics introduced in the past six years, focusing on their influence and effectiveness. We evaluated these algorithms across essential criteria: citation frequency, diversity in tackled problem types, code availability, ease of parameter tuning, introduction of novel mechanisms, and resilience to issues like stagnation and early convergence. Out of 158 algorithms, we identified 23 that set themselves apart, each contributing unique solutions to long-standing optimization challenges. These algorithms stand out for their versatility and innovation, positioning them as valuable assets for advancing research and addressing complex real-world problems. Our review offers a detailed analysis of these algorithms, comparing their strengths, limitations, similarities, and applications, while highlighting promising trends and future pathways in metaheuristic research.

\begin{figure}[h]
\begin{center}  
\includegraphics[width=0.7\textwidth]{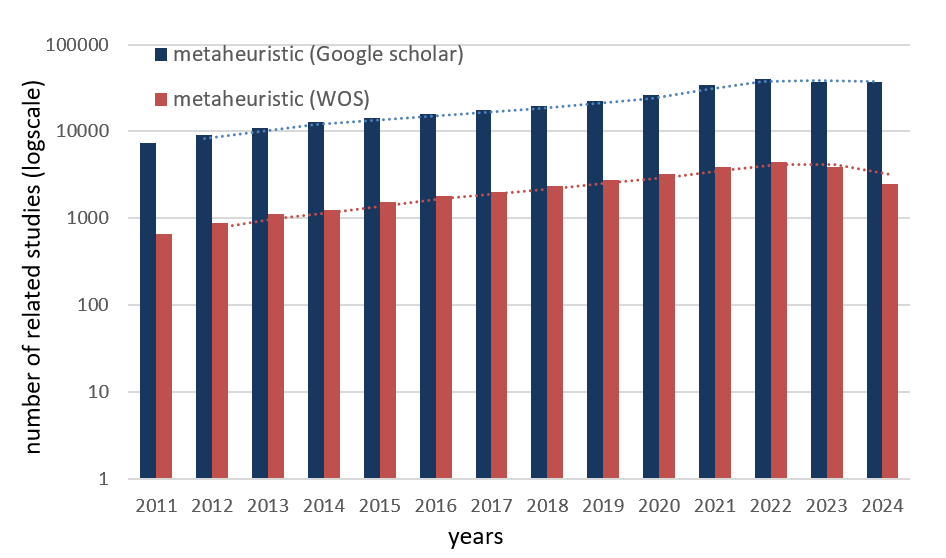}
\end{center} 
\caption{The query results on Google Scholar and Web of Science (WOS) with the keywords "metaheuristic" for the last 15 years (2011 and 2024).} \label{chart}
\end{figure}

Figure \ref{chart} presents query results from Google Scholar and Web of Science (WOS) using the keyword "metaheuristic" over the last 15 years (2011–2024). This graph highlights a steady, growing interest in the field, with new metaheuristics proposed each year and an increasing momentum in research activity. Such a sustained trend suggests that similar studies will likely continue as researchers explore innovative methods and variations within this domain. The ongoing introduction of new algorithms underscores a persistent need to tackle emerging optimization challenges. With applications spanning a wide range of areas, the demand for effective solutions is driving growth and innovation. Moving forward, advancements are anticipated to further enhance research outcomes, fostering improvements in both efficiency and effectiveness for complex problem-solving. Table \ref{metaheuristics} presents the most cited and influential metaheuristic algorithms selected for this review paper, showcasing those with substantial impact in the field. Table \ref{metaheuristics-2} introduces a second group of notable new metaheuristics. Beyond these, numerous additional articles on emerging metaheuristic methods were published within the same period but could not be included in our study.

\begin{center}
\begin{longtable}{llr}
\caption{Our selected list of the most influential metaheuristic algorithms developed between 2019 and 2024 (sorted by the number of citations received, as of November 2024)}\label{metaheuristics}\\
\hline
 \textbf{Metaheuristic} & \textbf{year} &  \textbf{\#citations} \\ 
\hline
 Harris hawks optimization \citep{heidari2019harris} 	&	2019	&	11300	\\
 Butterfly optimization algorithm \citep{arora2019butterfly} 	&	2019	&	6150	\\
 Gradient-based optimizer \citep{ahmadianfar2020gradient} 	&	2020	&	5990	\\
 Slime mould algorithm \citep{li2020slime} 	&	2020	&	5570	\\
 Marine predators algorithm \citep{faramarzi2020marine} 	&	2020	&	5080	\\
 Equilibrium optimizer \citep{faramarzi2020equilibrium} 	&	2020	&	4890	\\
 Aquila optimizer \citep{abualigah2021aquila} 	&	2021	&	3300	\\
 Seagull optimization \citep{dhiman2019seagull} 	&	2019	&	3050	\\
 Manta ray foraging optimization \citep{zhao2020manta} 	&	2020	&	2990	\\
 Chimp optimization algorithm \citep{khishe2020chimp} 	&	2020	&	2420	\\
 Squirrel Search Algorithm  \citep{jain2019novel} 	&	2019	&	2280	\\
 Henry gas solubility optimization \citep{hashim2019henry} 	&	2019	&	2150	\\
 Archimedes optimization algorithm \citep{hashim2021archimedes} 	&	2021	&	2080	\\
 Tunicate swarm algorithm \citep{kaur2020tunicate} 	&	2020	&	2020	\\
 Honey badger algorithm \citep{hashim2022honey} 	&	2022	&	1970	\\
 Mayfly optimization \citep{zervoudakis2020mayfly} 	&	2020	&	1720	\\
 African vultures optimization algorithm \citep{abdollahzadeh2021african} 	&	2021	&	1250	\\
 Golden jackal optimization \citep{chopra2022golden} 	&	2022	&	985	\\
 Dung beetle optimizer \citep{xue2023dung} 	&	2023	&	966	\\
 Coati Optimization Algorithm \citep{dehghani2023coati} 	&	2023	&	769	\\
 Chaos game optimization \citep{oueslati2024chaos} 	&	2024	&	767	\\
 Beluga whale optimization \citep{zhong2022beluga} 	&	2022	&	710	\\
 Gazelle optimization algorithm \citep{agushaka2023gazelle} 	&	2023	&	442	\\
\hline
\end{longtable}
\end{center}

No study like ours provides such a comprehensive examination of the metaheuristic algorithms, and no other highlights these 23 new algorithms with similar depth. Our analysis uniquely evaluates their impact and potential, offering insights that distinguish it from previous work in this field.

The contributions of our review {can be listed as follows}:
\begin{itemize} 
   \item Identification of 23 influential metaheuristic algorithms introduced between 2019 and 2024, based on criteria such as citation count, problem diversity, code availability, ease of parameter tuning, and resistance to optimization issues. 
   \item Detailed analysis of selected algorithms, examining unique mechanisms, strengths, and limitations, to guide researchers and practitioners in selecting suitable algorithms for diverse optimization challenges. 
   \item Evaluation of algorithm accessibility, parameter setting, binary encoding, and ease of implementation to encourage broader usability and adoption in academic and industrial contexts. 
   \item  Listing the state-of-the-art applications of metaheuristic algorithms between 2019 to 2024 in different domains.
   \item Synthesis of emerging trends within the metaheuristic field, including new mechanisms, hybrid models, and similar strategies, provides insights into ongoing research directions and future exploration areas. 
\end{itemize}

Section \ref{section2} presents an overview of surveys and reviews on metaheuristic algorithms conducted over the last six years (2019–2024). Section \ref{section3} summarizes the 23 most influential metaheuristics selected from the same period, detailing their mathematical formulations and briefly explaining a few related studies. Section \ref{section4} highlights recent applications of metaheuristic algorithms, with a primary focus on their usage from 2019 to 2024. Section \ref{section5} discusses current challenges associated with recent metaheuristic algorithms. The final section offers concluding remarks and suggests directions for future research.

\section{Previous surveys}\label{section2}

This section summarizes metaheuristic survey/review articles published between 2019 and 2024. \citet{hussain2019metaheuristic} reviewed 1222 publications from 1983 to 2016, addressing four key dimensions: new algorithms, modifications, comparisons, and future research gaps, with the objective of highlighting potential open questions and critical issues raised in the literature. The work provides guidance for future research to be conducted more meaningfully that can serve the advancement of this area of research. \citet{halim2021performance} studied simulation-driven metaheuristic algorithms that outperform deterministic ones in solving various problems, but their stochastic nature can result in varied solution quality. Accurate performance assessment requires appropriate measurement tools focusing on both efficiency—speed and convergence—and effectiveness—solution quality—while statistical analysis is crucial for evaluating effectiveness. \citet{wong2019review} provided an overview of evolutionary algorithms, focusing on three key areas: state-of-the-art algorithms, benchmarking issues, and recent successful applications, reflecting the significant growth in research and applications over the past two decades. Significant advancements in metaheuristic algorithms have been made since their inception, with numerous new algorithms emerging daily, highlighting the need to identify the best-performing ones for sustained application. \citet{dokeroglu2019survey} identified fourteen notable metaheuristics introduced between 2000 and 2020, chosen for their efficiency, high citation counts, and unique features, while also exploring recent research trends, hybrid approaches, theoretical gaps, and new opportunities in the field. \citet{agushaka2022initialisation} surveyed various initialization schemes aimed at improving the solution quality of population-based metaheuristic algorithms, emphasizing the importance of population size and diversity; it categorizes popular schemes—such as random numbers, quasirandom sequences, chaos theory, and hybrids—discusses their effectiveness and limitations, identifies research gaps, and compares the impact of ten initialization methods on the performance of three metaheuristic optimizers: the bat algorithm, Grey Wolf Optimizer, and butterfly optimization algorithm.

\citet{abd2021advanced} examined metaheuristic algorithms developed between 2014 and 2020, detailing their characteristics and the modifications that have been implemented.  \citet{agrawal2021metaheuristic} provided an extensive literature review of metaheuristic algorithms developed from 2009 to 2019 for feature selection, categorizing over a hundred algorithms based on their behavior and focusing specifically on binary variants. It details each algorithm's classification, the classifiers used, datasets, and evaluation metrics, while also identifying challenges and issues in obtaining optimal feature subsets. Additionally, it highlights research gaps for future work in developing or modifying metaheuristic algorithms for classification, concluding with a case study utilizing datasets from the UCI repository to demonstrate the application of various metaheuristic algorithms in achieving optimal feature selection. \citet{talbi2021machine} aim to fill the gap in comprehensive surveys and taxonomies on this topic by exploring various synergies between ML and metaheuristics, proposing a detailed taxonomy based on search components and target optimization problems. Additionally, it seeks to inspire researchers in optimization to integrate ML concepts into metaheuristics while identifying open research issues that warrant further exploration. 

\citet{rajwar2023exhaustive} studied reviews approximately 540 metaheuristics, providing statistical insights and addressing the prevalence of similarities among algorithms with different names, raising the question of whether modifications qualify as "novel." It introduces a new taxonomy based on the number of control parameters, highlights real-world applications of metaheuristics, and identifies limitations and challenges that could inform future research directions. While much progress has been made, many unexplored areas remain, making this study a valuable resource for newcomers and the broader research community. \citet{osaba2021tutorial} proposed a set of good practices to enhance scientific rigour, value, and transparency in metaheuristic research, introducing a comprehensive methodology that guides researchers through each phase of their studies. Key aspects—including problem formulation, solution encoding, implementation of search operators, evaluation metrics, experiment design, and real-world performance considerations—will be discussed, along with challenges and future research directions necessary for successfully deploying new optimization metaheuristics in real-world applications. \cite{dokeroglu2022comprehensive} highlighted the most effective recent metaheuristic feature selection algorithms, focusing on their exploration/exploitation operators, selection methods, transfer functions, fitness evaluations, and parameter settings. It also addresses current challenges faced by these algorithms and suggests future research topics for further exploration in the field.

\citet{gharehchopogh2023quantum} presented an overview of various applications of quantum computing in metaheuristics, offering a classification of quantum-inspired metaheuristic algorithms for optimization problems. The main aim of this paper is to summarize and discuss the applications of these algorithms across science and engineering, highlighting their potential and effectiveness in solving complex optimization challenges. \citet{abualigah2022meta} presented the results of state-of-the-art optimization methods to identify which versions perform best in addressing specific problems. It highlights significant future research directions for potential methods, covering key topics in engineering and artificial intelligence. By compiling a substantial number of published works on metaheuristic optimization methods applied to various engineering design problems, this review serves as a valuable resource for future researchers exploring the intersection of metaheuristics and engineering design.

\citet{ezugwu2021metaheuristics} introduced a new taxonomic classification of both classical and contemporary metaheuristic algorithms, aiming to provide an easily accessible collection of popular optimization tools for the global optimization research community tackling complex real-world problems. Additionally, a bibliometric analysis of the field of metaheuristics over the past 30 years is included, offering insights into research trends and developments in this area. The application of computational intelligence and soft computing techniques is essential for addressing multi-objective problems and managing trade-offs among control performance objectives. \citet{rodriguez2020multi} reviewed the literature on multi-objective metaheuristics used in intelligent control, specifically focusing on controller tuning problems, and discusses their effectiveness in solving complex challenges while maintaining reasonable computational costs.

Parameter tuning is essential for optimizing algorithm performance in metaheuristics, and automating this process has gained significant attention in recent years. \citet{huang2019survey}  provided a comprehensive survey of automatic parameter tuning methods, introducing a new taxonomy that categorizes them into simple, iterative, and high-level generate-evaluate methods, while discussing their strengths, weaknesses, and future research directions. The Resource-Constrained Project Scheduling Problem (RCPSP) is a well-known NP-hard problem with applications in manufacturing, project management, and production planning, primarily addressed through heuristic methods. \cite{pellerin2020survey} surveyed the evolution of hybrid metaheuristic approaches developed over the last two decades to solve the RCPSP, providing descriptions of the fundamental principles behind these hybrids, comparing their results on PSPLIB data instances, and discussing the distinguishing features of the most effective hybrid methods.

\citet{elshaer2020taxonomic} built on a previous taxonomic review of the Vehicle Routing Problem (VRP) literature by classifying VRP and its variants solved using metaheuristic algorithms and investigating the contributions of each algorithm from 2009 to 2017. By analyzing 299 articles, the study reveals trends in algorithm usage and identifies popular VRP variants, highlighting promising topics for future research. \citet{essaid2019gpu} This survey explored the use of parallel computing, particularly GPU-based implementations, to enhance execution speed and solution quality, presenting mechanisms for GPU programming in parallel metaheuristics and discussing findings from relevant research studies. \citet{hussain2019exploration} conducted an in-depth empirical analysis of five swarm-based metaheuristic algorithms, quantitatively examining their exploration and exploitation, revealing that coherence and consistency among swarm individuals are crucial for success, and suggesting that this analytical approach can be used for component-wise diversity analysis to enhance search strategies. \citet{mohammed2019systematic} presented a systematic meta-analysis of the whale optimization algorithm (WOA), detailing its algorithmic background, characteristics, limitations, and applications, while highlighting its superior convergence speed and balance between exploration and exploitation compared to other optimization algorithms. Additionally, it introduces a hybrid approach combining WOA with the BAT algorithm, demonstrating that the WOA-BAT hybrid outperforms WOA in 16 benchmark functions and excels in various challenges from CEC2005 and CEC2019.

\citet{kareem2022metaheuristic} examined, compared, and described various metaheuristic algorithms, including Genetic Algorithm (GA), Ant Colony Optimization (ACO), Simulated Annealing (SA), Particle Swarm Optimization (PSO), and Differential Evolution (DE). It concludes by presenting the performance results of each algorithm across different environments. \citet{abd2021advanced} provided a comprehensive survey of recent optimization methods, specifically swarm intelligence (SI) and evolutionary computing (EC), used to enhance DNN performance, analyzing their role in optimizing hyperparameters and structures for handling massive-scale data, while also identifying potential directions for future improvements and open challenges in evolutionary DNNs.

\section{The most influential metaheuristics (between 2019 and 2024)}\label{section3}

In this section, we provide a comprehensive overview of the latest advancements in metaheuristic algorithms developed between 2019 and 2024, showcasing key examples of their state-of-the-art applications. Our selection criteria included citation impact, solution efficacy, adaptability across diverse domains, and innovations in exploration-exploitation techniques as well as mechanisms for managing local optima. We anticipate that these metaheuristics will gain greater prominence and see more widespread application in the coming years, distinguishing them from other recent algorithms in the field.

\subsection{Harris Hawk Optimization}

\begin{figure}[h]
\begin{center}  
\includegraphics[width=0.65\textwidth]{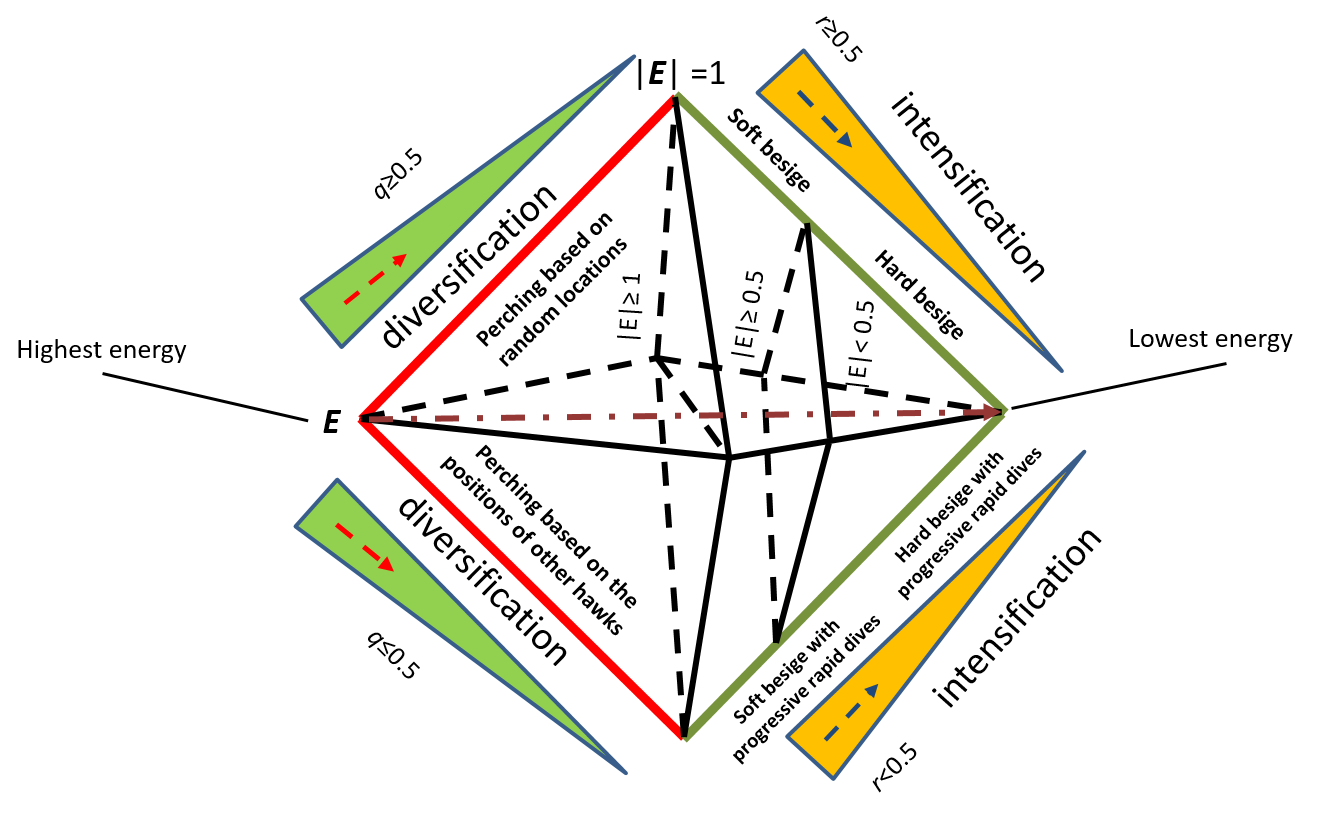}
\end{center} 
\caption{The steps of the HHO metaheuristic according to the energy level (E), $q$ and $r$ values.} \label{Harris}
\end{figure}

The Harris Hawk Optimization (HHO) algorithm is a powerful metaheuristic inspired by the cooperative hunting strategies of Harris Hawks, known for their group-based tactics and sudden attacks on prey \citep{heidari2019harris,alabool2021harris}. The HHO algorithm is particularly effective in solving complex optimization problems characterized by non-linearity, high dimensionality, and multiple local optima. HHO simulates hawks scouting for prey and executing coordinated surprise pounces, enabling both exploration (broad search) and exploitation (local search) of the solution space. See Figure \ref{Harris} for the steps of the HHO according to the energy level (E), $q$ and $r$ values.

HHO starts with a random population of candidate solutions, which are evaluated and iteratively updated based on dynamic behaviors that mimic real hawk hunting. During the exploration phase, the algorithm searches broadly to avoid local optima. If a promising area is identified, the exploitation phase begins, where hawks perform strategic, sudden moves to converge quickly on high-quality solutions. These moves are influenced by adaptive parameters that mimic the prey’s escape patterns, helping balance between global and local searches.

Initialize a population of $N$ hawks, represented by:
\begin{equation}
X_i = (x_{i,1}, x_{i,2}, \ldots, x_{i,d}), \quad i = 1, 2, \ldots, N,
\end{equation}
where $d$ is the dimension of the search space. The positions of the hawks are initialized randomly within the problem boundaries. During the exploration phase, Hawks randomly search for prey based on their current position and a reference leader (best solution found so far):
\begin{equation}
X_i(t+1) = X_{\text{rand}}(t) - r_1 |X_{\text{rand}}(t) - 2r_2 X_i(t)|,
\end{equation}
where $X_{\text{rand}}$ is a randomly chosen hawk, $r_1$ and $r_2$ are random numbers uniformly distributed in $[0, 1]$.

The transition from exploration to exploitation depends on the prey's behavior and escape energy $E$:
\begin{equation}
E = 2 E_0 (1 - \frac{t}{T}),
\end{equation}
where $E_0$ is a random number in $[-1, 1]$, $t$ is the current iteration, and $T$ is the maximum number of iterations.

If $|E| \geq 0.5$, the hawks perform the soft besiege (Exploitation Phase):
\begin{equation}
X_i(t+1) = \Delta X(t) - E |J X_{\text{best}}(t) - X_i(t)|,
\end{equation}
where $\Delta X(t)$ is the difference between the best and current solutions, and $J$ is a random jump strength coefficient.

If $|E| < 0.5$, the hawks perform the hard besiege:
\begin{equation}
X_i(t+1) = X_{\text{best}}(t) - E |X_{\text{best}}(t) - X_i(t)|.
\end{equation}

In case of random attacks or surprise pounces, hawks simulate abrupt dives:
\begin{equation}
X_i(t+1) = X_{\text{prey}}(t) - E (|X_{\text{prey}}(t) - X_i(t)|^{\beta}),
\end{equation}
where $\beta$ is a control parameter that simulates the sudden movements.


\citet{kamboj2020intensify} enhanced the global search capabilities and prevented local optima, a hybrid variant called the HHO-Sine Cosine Algorithm (hHHO-SCA). This variant integrates the Sine-Cosine Algorithm (SCA) exploration mechanisms into the HHO to improve its performance. The hHHO-SCA has been tested on complex, nonlinear, non-convex, and highly constrained engineering design problems. Results demonstrated that hHHO-SCA outperformed the standard SCA, HHO, and other optimization algorithms like Ant Lion Optimizer, Moth-Flame Optimization, and Grey Wolf Optimizer. The proposed algorithm showed superior performance across diverse optimization problems, supporting its effectiveness in solving multidisciplinary design and engineering tasks. \citet{elgamal2020improved} introduced CHHO that has two significant enhancements to the standard HHO. First, chaotic maps are applied during the initialization phase to improve population diversity, allowing better search space exploration. Second, Simulated Annealing (SA) is integrated to refine the current best solution, boosting the algorithm’s exploitation capabilities. \citet{too2019new} proposed Quadratic Binary HHO (QBHHO) that aims to improve the exploration and exploitation balance, providing better solutions for feature selection. The effectiveness of BHHO and QBHHO was validated using 22 datasets from the UCI machine learning repository. 
\subsection{Butterfly optimization algorithm}

\begin{figure}[h]
\begin{center}  
\includegraphics[width=0.55\textwidth]{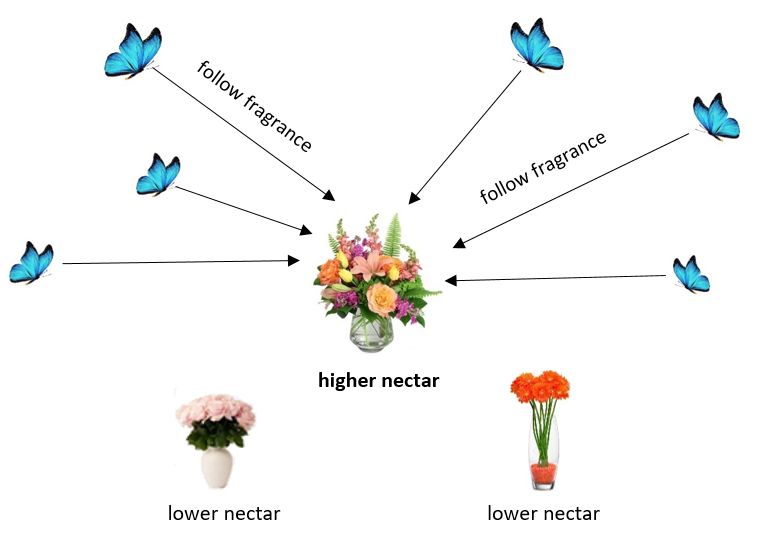}
\end{center} 
\caption{The movement behavior of butterflies} \label{butterfly}
\end{figure}

The Butterfly Optimization Algorithm (BOA) is a metaheuristic based on the foraging and mating behavior of butterflies \citep{arora2019butterfly}. It simulates the way butterflies use their sense of smell to find food or mates, balancing exploration and exploitation in the search space. BOA has been successfully applied to various optimization problems, demonstrating competitive performance in finding optimal solutions (See Figure \ref{butterfly} for the movement behavior of butterflies). 

Butterflies perceive the quality of a solution (fitness) via sensory perception modeled as fragrance. The fragrance is defined as:

\begin{equation}
f_i = c \cdot I_i^a
\end{equation}

where $f_i$ is the fragrance of butterfly $i$, $c$ is a constant,  $I_i$ is the fitness of the solution (smell intensity), and $a$ is a sensory modality parameter, controlling the degree of perception. The movement of butterflies is controlled by both global and local search strategies, depending on the fragrance perceived. The global search allows butterflies to move towards the best solution in the population:

\begin{equation}
X_i(t+1) = X_i(t) + r \cdot f_i \cdot (X_{\text{best}} - X_i(t)) 
\end{equation}

where $X_{\text{best}}$ is the best solution found so far, $r$ is a random number in $[0,1]$, and $f_i$ is the fragrance of butterfly $i$. For local search, the movement is determined by the fragrance of nearby butterflies:

\begin{equation}
X_i(t+1) = X_i(t) + r \cdot f_i \cdot (X_j(t) - X_k(t))
\end{equation}

where $X_j(t)$ and $X_k(t)$ are two randomly selected butterflies. To switch between global and local search, a random switching probability $p$ is introduced:

\begin{equation}
p = \text{rand}(0,1)
\end{equation}

If $p$ is less than a threshold $p_0$, a global search is performed; otherwise, local search is executed. This mechanism ensures a balance between exploration and exploitation. The sensory modality parameter $a$ is adapted over iterations to fine-tune the algorithm:

\begin{equation}
a(t) = a_{\text{min}} + (a_{\text{max}} - a_{\text{min}}) \cdot \frac{t}{T}
\end{equation}

where $a_{\text{min}}$ and $a_{\text{max}}$ define the range for the sensory modality, $t$ is the current iteration, and $T$ is the total number of iterations.

\citet{tubishat2020dynamic} introduces Dynamic BOA (DBOA), addressing its limitations in high-dimensional problems, such as local optima stagnation and lack of solution diversity. By incorporating a Local Search Algorithm Based on Mutation (LSAM), DBOA improves solution diversity and avoids local optima. Experiments on 20 UCI benchmark datasets show that DBOA outperforms other algorithms across various performance metrics. \citet{makhadmeh2023recent} introduced information about the BOA to illustrate the essential foundation and its relevant optimization concepts. In addition, the BOA inspiration and its mathematical model are provided with an illustrative example to prove its high capabilities. Subsequently, all reviewed studies are classified into three main classes based on the adaptation form, including original, modified, and hybridized. The main BOA applications are also thoroughly explained. Furthermore, the BOA advantages and drawbacks in dealing with optimization problems are analyzed. Finally, the paper is summarized in conclusion with the future directions that can be investigated further. \citet{alweshah2022monarch} applied the monarch BOA (MBO) algorithm with a wrapper FS method using the KNN classifier. Tested on 18 benchmark datasets, MBO outperformed four metaheuristic algorithms (WOASAT, ALO, GA, and PSO), achieving an average classification accuracy of 93\% and significantly reducing the feature selection size. The results demonstrate MBO's effectiveness and efficiency in FS, with a strong balance between global and local search.

\subsection{Gradient-based optimizer}

Gradient-based optimizer (GBO) combines the gradient and population-based methods, the search direction is specified by the Newton’s method to explore the search domain utilizing a set of vectors and two main operators (i.e., gradient search rule and local escaping operators). Minimization of the objective function is considered in the optimization problems \citep{ahmadianfar2020gradient,daoud2023gradient}. Gradient-based optimizers operate by calculating gradients—essentially the slope or rate of change of the function with respect to the model parameters—and then updating these parameters in the direction that reduces the objective function, aiming for an optimal or near-optimal solution.

One of the most widely used gradient-based optimizers is Stochastic Gradient Descent (SGD) \citep{amari1993backpropagation}, which updates parameters based on the gradient calculated for a single or mini-batch of samples. This approach is faster than full-batch gradient descent \citep{hinton2012neural}, particularly for large datasets, but can suffer from noisy updates and may struggle with complex optimization landscapes. To address these issues, variants like Momentum, Nesterov Accelerated Gradient, Adagrad, RMSprop, and Adam (Adaptive Moment Estimation) have been developed.

Gradient-based optimizers are essential in fields like deep learning, reinforcement learning, and computer vision, where they enable efficient training of large models by focusing on regions in parameter space that progressively reduce error. However, their reliance on gradients also makes them susceptible to challenges like getting trapped in local minima or saddle points, especially in high-dimensional non-convex problems. Consequently, researchers are continuously developing enhancements and alternative algorithms to make these optimizers more robust across various machine learning applications.


\citet{premkumar2021mogbo} introduced a multiobjective GBO (MOGBO), for solving multiobjective truss-bar design problems. MOGBO employs a gradient-based approach with operators like the local escaping operator and gradient search rule, using non-dominated sorting and crowding distance mechanisms to achieve Pareto optimal solutions. Performance tests on various benchmark problems show MOGBO outperforms other algorithms in accuracy, runtime, and metrics like hyper-volume and diversity, proving its effectiveness in complex multiobjective optimization tasks. \citet{jiang2021efficient} proposed eight variants of the binary GBO utilizing S-shaped and V-shaped transfer functions to convert the search space to a discrete format. The performance of these binary GBO algorithms is evaluated on 18 UCI datasets and 10 high-dimensional datasets, comparing them against other feature selection methods. Results indicate that one binary GBO variant outperforms other algorithms, demonstrating superior overall performance in various metrics. \citet{helmi2021novel} introduced a new algorithm (GBOGWO), a feature selection method that enhances the GBO with Grey Wolf Optimizer (GWO) operators, to address high-dimensional data challenges and improve HAR classification. Using UCI-HAR (Human Activity Recognition) and WISDM datasets, GBOGWO achieved an average classification accuracy of 98\%, demonstrating its effectiveness in refining HAR model performance.
\subsection{Slime mould algorithm}

\begin{figure}[h]
\begin{center}  
\includegraphics[width=0.4\textwidth]{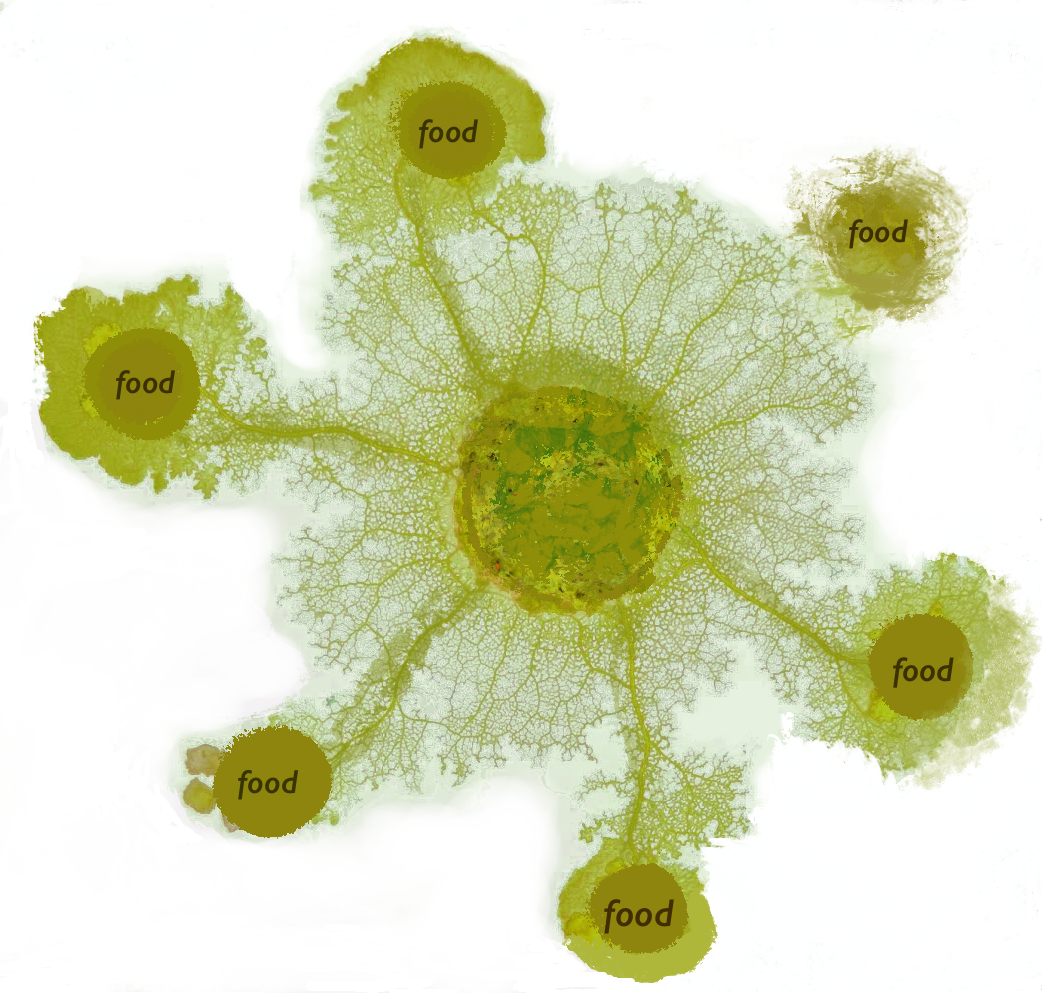}
\end{center} 
\caption{Foraging morphology of slime mould} \label{Slime}
\end{figure}

The Slime Mould Algorithm (SMA) primarily simulates the behavior and morphological changes of the slime mould Physarum polycephalum during its foraging process, rather than modeling its entire life cycle \citep{li2020slime,chen2023slime}. This organism is a eukaryote that thrives in cold, humid environments. The primary nutritional stage is the plasmodium, which represents the active and dynamic phase of the slime mould and is the focal point of this survey. During this phase, slime mould actively searches for food, encircles it, and releases enzymes for digestion. As it migrates, the leading edge expands into a fan shape, supported by a network of interconnected veins that facilitate the flow of cytoplasm, as illustrated in Fig. \ref{Slime}. Due to their unique structure and behavior, slime moulds can simultaneously utilize multiple food sources, forming a network that connects them. See Figure \ref{Slime} for the foraging morphology of slime mould.

The slime mould is capable of locating food sources by detecting odours in the air. To mathematically model this foraging behavior, the following formula have been proposed to simulate the contraction mode. The slime mould can navigate towards food sources by detecting odours in the air. To mathematically represent this approaching behavior, the following formulas have been proposed to simulate the contraction mode:

\begin{equation}
\vec{X(t + 1)} =
\begin{cases}
\vec{X_b(t)} + \vec{vb} \cdot \left( \vec{W} \cdot \left( \vec{X_A(t)} - \vec{X_B(t)} \right) \right), & r < p \\
\vec{vc} \cdot \vec{X(t)}, & r \geq p
\end{cases}
\label{sma_app_food}
\end{equation}
where \(\vec{vb}\) is a parameter that ranges from \([-a, a]\), and \(\vec{vc} \) decreases linearly from one to zero. The variable \(t\) denotes the current iteration, \(\vec{X_b}\) indicates the location of the individual with the highest odour concentration detected, \(\vec{X}\) represents the position of the slime mould, and \(\vec{X_A}\) and \(\vec{X_B}\) are two individuals randomly selected from the slime mould population. Additionally, \(\vec{W}\) signifies the weight of the slime mould which is formulated as follows:
\begin{equation}
\vec{W(\text{SmellIndex}(i))} =
\begin{cases}
1 + r \cdot \log \left( \frac{bF - S(i)}{bF - wF} + 1 \right), & \text{condition} \\
1 - r \cdot \log \left( \frac{bF - S(i)}{bF - wF} + 1 \right), & \text{otherwise}
\end{cases}
\label{sma_app_food_w}
\end{equation}

where "condition" indicates that \(S(i)\) ranks in the top half of the population, \(r\) represents a random value within the interval \([0, 1]\), \(bF\) signifies the best fitness value achieved during the current iteration, \(wF\) represents the worst fitness value obtained thus far in the iterative process, and \(\text{SmellIndex}\) refers to the sequence of fitness values sorted in ascending order for the minimum value problem.

The position of the searching individual \(\vec{X}\) can be updated based on the best location \(\vec{X}_b\) currently identified, and the adjustment of parameters \(\vec{v}_b\), \(\vec{v}_c\), and \(\vec{W}\) can modify the individual's location. The inclusion of the random variable in the formula allows individuals to create search vectors at any angle, enabling them to explore the solution space in all directions, which enhances the algorithm's potential for finding the optimal solution. 


The next step is the contraction mode of the venous tissue structure of slime mould when searching. The greater the concentration of food encountered by the vein, the stronger the wave produced by the bio-oscillator, resulting in a faster flow of cytoplasm and a thicker vein. Equation \ref{sma_app_food_w} mathematically models the positive and negative feedback between the vein width of the slime mould and the food concentration that was investigated. The component \(r\) in Equation \ref{sma_app_food_w} represents the uncertainty in the mode of venous contraction. The logarithm is utilized to moderate the rate of change in numerical values, ensuring that the contraction frequency does not fluctuate excessively. The "condition" reflects how the slime mould adjusts its search patterns based on food quality. When food concentration is high, the weight in that area increases; conversely, when food concentration is low, the weight diminishes, prompting the slime mould to explore new regions. Based on the aforementioned principles, the mathematical formula for updating the location of the slime mould is as follows:
\begin{equation}
\vec{X^{*}} =
\begin{cases}
\text{rand} \cdot (UB - LB) + LB, & \text{if } \text{rand} < z \\
\vec{X_b(t)} + \vec{vb} \cdot \left( W \cdot \vec{X_A(t)} - \vec{X_B(t)} \right), & \text{if } r < p \\
\vec{vc} \cdot \vec{X(t)}, & \text{if } r \geq p
\end{cases}
\end{equation}
where \(LB\) and \(UB\) represent the lower and upper boundaries of the search range, respectively, and \(\text{rand}\) and \(r\) signify random values within the interval \([0, 1]\), \(z\) is used for oscillation.


\citet{chen2023slime} studied and analyzed key research related to the development of the SMA. A total of 98 SMA-related studies were retrieved, selected, and identified from the Web of Science database. The review focuses on two main aspects: advanced versions of the SMA and its application domains.  \citet{premkumar2020mosma} presented a Multi-objective SMA  (MOSMA) for tackling multi-objective optimization challenges in industrial settings, based on the oscillatory behaviors of slime mould in laboratory experiments. MOSMA integrates the core principles of SMA with elitist non-dominated sorting and a crowding distance operator to ensure broad coverage of Pareto optimal solutions. Tested across 41 diverse case studies, MOSMA outperformed existing algorithms (MOSOS, MOEA/D, MOWCA) on several performance metrics, demonstrating its strong capability for handling complex multi-objective optimization problems. \citet{houssein2022efficient}  developed a multi-objective SMA, called MOSMA, for solving complex multi-objective optimization problems. MOSMA incorporates an external archive to store and manage Pareto optimal solutions, simulating the social behaviors of slime mould in a multi-objective search space. Validated on CEC’20 benchmarks and various engineering problems, MOSMA outperforms six established algorithms (e.g., MOGWO, NSGA-II) in terms of solution proximity to the Pareto set and inverted generational distance, proving its strength in real-world applications like automotive helical coil spring optimization.

\subsection{Marine Predators Algorithm}

\begin{figure}[h]
\begin{center}  
\includegraphics[width=0.6\textwidth]{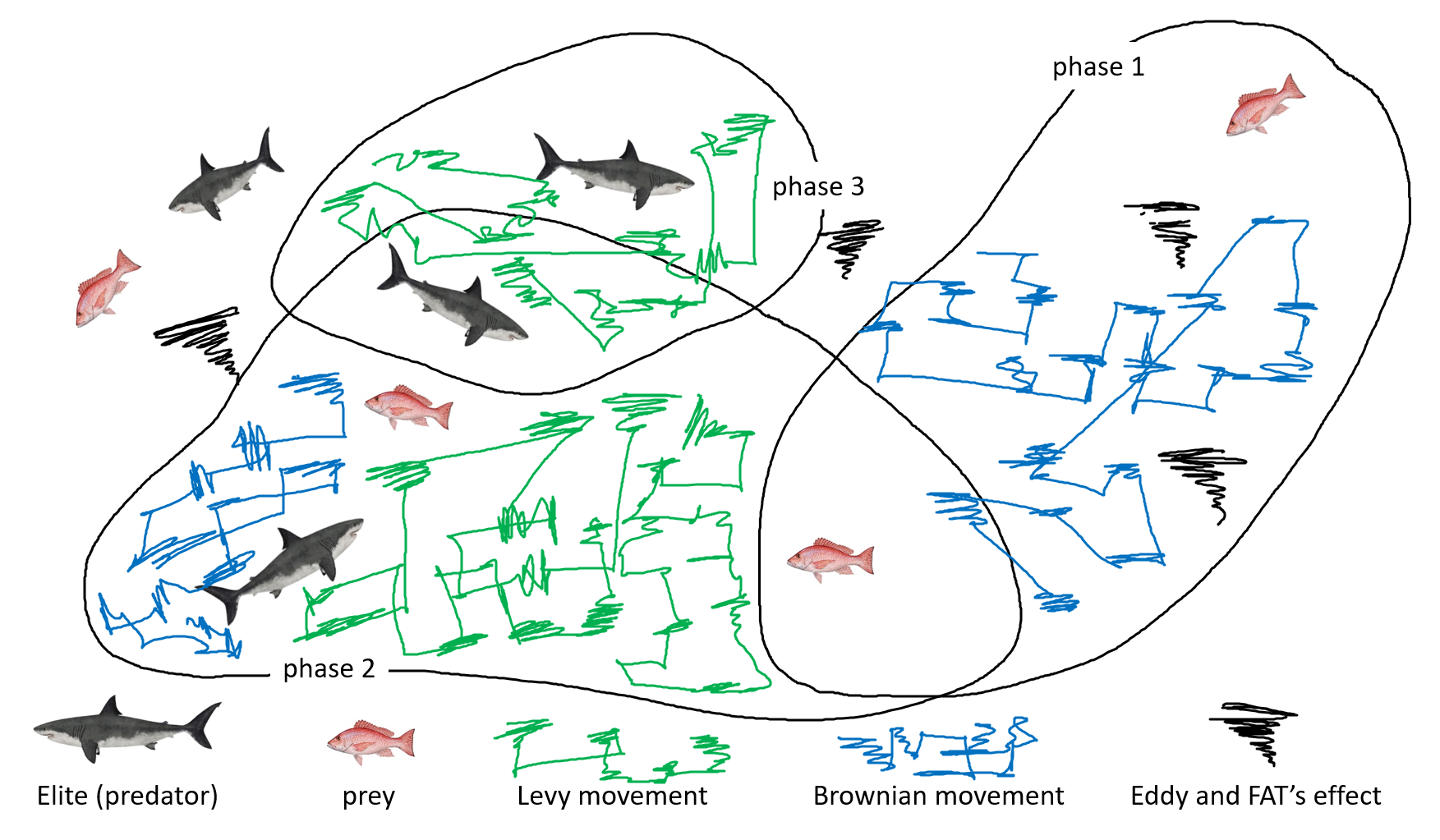}
\end{center} 
\caption{Three phases of the marine predators metaheuristic} \label{marine}
\end{figure}

The Marine Predators Algorithm (MPA) simulates the behavior of marine predators foraging in the ocean \citep{faramarzi2020marine}. The algorithm primarily relies on different movement phases that represent various predation strategies based on the interaction between predators and prey. See Figure \ref{marine} for the phases of the MPA.

In the initial exploration phase, the algorithm uses random movements inspired by Lévy flights. The position of each predator \(X_i\) at iteration \(t+1\) is updated as follows:
\[
\mathbf{X}_i^{t+1} = \mathbf{X}_i^t + r \cdot \text{Lévy}(\lambda) \cdot (\mathbf{X}_i^t - \mathbf{X}_{\text{best}}^t),
\]
where \(r\) is a random number in the range \([0,1]\), \(\text{Lévy}(\lambda)\) represents a Lévy flight with scaling parameter \(\lambda\), and \(\mathbf{X}_{\text{best}}^t\) is the position of the best solution found so far.

The exploitation phase adjusts the movement based on a "Brownian motion" mechanism if the prey is close to the predator. The position update in this phase is given by:
\[
\mathbf{X}_i^{t+1} = \mathbf{X}_{\text{best}}^t + B \cdot (\mathbf{X}_{i}^t - \mathbf{X}_{\text{best}}^t),
\]
where \(B\) represents a random Brownian motion.

Alternatively, if the prey is farther away, the algorithm uses a different movement strategy:
\[
\mathbf{X}_i^{t+1} = \mathbf{X}_{\text{best}}^t + F \cdot (\mathbf{X}_i^t - \mathbf{X}_{\text{mean}}^t),
\]
where \(F\) is a random factor, and \(\mathbf{X}_{\text{mean}}^t\) is the mean position of all solutions at iteration \(t\).

To model the escape of prey, a diversification strategy is applied:
\[
\mathbf{X}_i^{t+1} = \mathbf{X}_i^t + S \cdot (\mathbf{X}_i^t - \mathbf{X}_{\text{worst}}^t),
\]
where \(S\) is a scaling factor, and \(\mathbf{X}_{\text{worst}}^t\) is the position of the worst solution.


\citet{abd2021efficient}  introduced MPA-KNN, a novel hybridization of the MPA and k-Nearest Neighbors (k-NN), to improve feature selection for medical datasets, with feature sizes ranging from tiny to massive. Experimental results show that MPA-KNN outperforms eight well-regarded metaheuristic algorithms in accuracy, sensitivity, and specificity across 18 UCI medical benchmarks, underscoring its effectiveness for optimal feature selection. \citet{ramezani2021new}  proposed an enhanced MPA variant that integrates opposition-based learning, chaotic mapping, self-adaptive population techniques, and an adaptive phase-switching mechanism for improved exploration and exploitation. Simulations conducted on CEC-06 2019 test functions and a real-world control problem applied to a DC motor indicate that the improved algorithm significantly outperforms the original MPA and five other optimization algorithms in accuracy and robustness. \citet{abdel2021parameter} presented an enhanced MPA for optimized photovoltaic parameter extraction, incorporating a population improvement strategy where adaptive mutation enhances high-quality solutions, and low-quality solutions are updated based on the best and high-ranked solutions. Experimental results demonstrate that the proposed algorithm offers superior accuracy, showing a high correlation with measured current-voltage data and proving effective for parameter estimation.
\subsection{Equilibrium optimizer} 

The Equilibrium Optimizer (EO) is inspired by dynamic mass balance models used in control systems, where a system reaches equilibrium \citep{faramarzi2020equilibrium,makhadmeh2022hybrid}. EO mimics the process of reaching equilibrium through iterations, balancing exploration and exploitation using the control mechanism of concentration updating. The algorithm leverages different equilibrium candidates and adaptive control to guide the search process. The algorithm maintains an equilibrium pool consisting of multiple equilibrium candidates. The update for each individual's position towards these candidates is given by:

\begin{equation}
X_i(t+1) = X_i(t) + \lambda \cdot (X_{\text{eq}}(t) - X_i(t)) + F \cdot (X_{\text{eq}}(t) - X_{\text{rand}}(t))
\end{equation}

where $X_{\text{eq}}(t)$ is the position of the equilibrium candidate at iteration $t$, $\lambda$ is the random control parameter for exploration, $F$ is the control parameter for exploitation, and $X_{\text{rand}}(t)$ is a random solution to introduce diversity. The parameters $\lambda$ and $F$ are updated dynamically over time to balance exploration and exploitation:

\begin{equation}
\lambda = 1 - \frac{t}{T}
\end{equation}

\begin{equation}
F = \text{rand}(0,1)
\end{equation}

where: $t$ is the current iteration, and $T$ is the maximum number of iterations. The equilibrium candidates in the pool are updated to reflect the best solutions found so far. This ensures that individuals are attracted towards high-quality solutions while maintaining diversity:

\begin{equation}
X_{\text{eq}}(t+1) = X_{\text{best}}(t) + \beta \cdot (X_{\text{best}}(t) - X_{\text{mean}}(t))
\end{equation}

Where: $X_{\text{best}}(t)$ is the best solution at iteration $t$, $X_{\text{mean}}(t)$ is the mean solution across the population, and $\beta$ is a constant controlling the influence of the best solution. As iterations proceed, the control parameters $\lambda$ and $F$ help the algorithm converge towards the equilibrium by reducing random fluctuations and encouraging exploitation.

\citet{wang2021photovoltaic} proposed an improved EO using a neural network to enrich photovoltaic cell data, enhancing optimization efficiency. Tested on three diode models, it outperforms other algorithms, achieving lower error rates and improving both precision and reliability, making it highly effective for photovoltaic cell parameter estimation. \citet{abdel2020solar} presented an improved IEO that integrates linear reduction diversity (LRD) and local minima elimination (MEM) to enhance solution accuracy and convergence. By directing poor fitness particles toward optimal solutions, LRD accelerates convergence, while MEM reduces entrapment risks. Extensive tests on photovoltaic models demonstrate IEO's competitive performance, showing superior optimization for solar cell applications. \citet{gao2020efficient} introduced two binary EO (BEO) for feature selection, designed for classification tasks. The first maps the continuous EO into discrete forms using S-shaped and V-shaped transfer functions (BEO-S and BEO-V), while the second (BEO-T) uses the current optimal position. Tests on 19 UCI datasets show BEO-V2 outperforms other methods significantly.
\subsection{Aquila Optimizer}

\begin{figure}[h]
\begin{center}  
\includegraphics[width=0.55\textwidth]{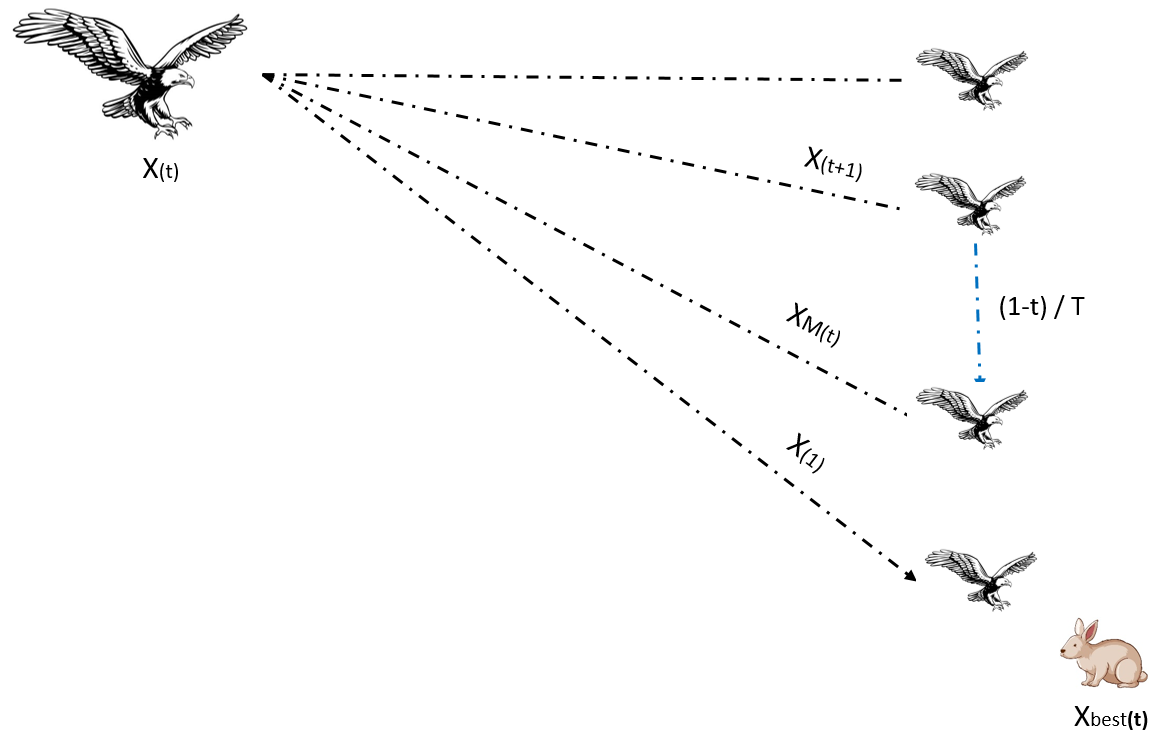}
\end{center} 
\caption{Aquila's high soar and vertical dive behavior} \label{Aquila}
\end{figure}

The Aquila Optimizer (AO) is inspired by the hunting behavior of Aquila, a bird of prey \cite{sasmal2023comprehensive}. AO mimics Aquila's powerful and efficient hunting strategies, combining exploration and exploitation to search for global optima. The algorithm consists of different movement strategies, which are applied dynamically to balance exploration of the search space and exploitation of promising regions (see Figure \ref{Aquila} for its soar and vertical dive behavior). 

The initial population of Aquilas is randomly generated. AO dynamically switches between different movement strategies depending on the stage of the hunt:

\begin{equation}
X_i(t+1) = X_{\text{best}}(t) + \alpha \cdot F(X_{\text{best}}(t) - X_i(t))
\end{equation}

During the hunting phase, the distance between the Aquila and the prey is calculated, influencing its strategy:

\begin{equation}
D = |C \cdot X_{\text{best}}(t) - X_i(t)|
\end{equation}

where $D$ is the distance between Aquila $i$ and the prey, $C$ is a coefficient representing the influence of the prey's position. In certain situations, Aquilas perform a dive to capture the prey with more precision, expressed as:

\begin{equation}
X_i(t+1) = X_{\text{best}}(t) + D \cdot e^{b \cdot l} \cdot \sin(2\pi l)
\end{equation}

where $b$ controls the width of the dive, $l$ is a random variable controlling the angle of the attack and $e$ is the base of the natural logarithm, indicating the sharpness of the dive. AO uses adaptive parameters to adjust the search dynamically. For example, $\alpha$ changes with time to balance between exploration and exploitation:

\begin{equation}
\alpha = 2 \cdot (1 - \frac{t}{T}) \cdot \text{rand}(0,1)
\end{equation}

where $T$ is the total number of iterations, and $t$ is the current iteration number.

\citet{al2022modified} addressed the shortcomings of the Adaptive Neuro-Fuzzy Inference System  (ANFIS) model in oil production estimation by optimizing its parameters with a modified AO and Opposition-Based Learning (OBL). The proposed model outperforms several modified ANFIS models and time-series forecasting methods using real-world datasets and performance metrics like Root Mean Square Error (RMSE) and Mean Absolute Error (MAE). \citet{mahajan2022hybrid} introduced a hybrid optimization method combining AO and Arithmetic Optimization Algorithm (AOA) to enhance convergence and solution quality. The proposed AO–AOA approach is evaluated on various problems, including image processing and engineering design, with consistent performance across both high- and low-dimensional problems. Population-based methods prove effective for high-dimensional optimization. \citet{bacs2023binary}  introduced the Binary AO (BAO) to address binary optimization problems, updating the continuous-based AO. The BAO uses transfer functions to convert the continuous search space into a binary one, with two variations: BAO-T and BAO-CM, which incorporate crossover and mutation steps. Tested on 63 knapsack problem datasets, BAO-CM outperformed BAO-T and other recent heuristic algorithms, demonstrating its effectiveness for binary optimization tasks.
\subsection{Seagull Optimization}

\citet{dhiman2019seagull} introduced the Seagull Optimization Algorithm (SOA), a bio-inspired approach based on seagull migration and attack behaviors to enhance exploration and exploitation within a search space. The SOA’s performance is benchmarked against nine popular metaheuristics across forty-four test functions, with evaluations of its computational complexity and convergence behavior. Additionally, SOA is applied to seven constrained real-world industrial problems, showcasing its effectiveness in addressing large-scale, complex optimization challenges. Experimental results demonstrate that SOA is highly competitive and well-suited for solving constrained, computationally expensive problems.


Seagulls' behavior can be described as follows: (i) During migration, seagulls travel in groups, starting from different positions to prevent collisions with one another, (ii) within the group, seagulls orient their movement toward the most fit individual, defined as the seagull with the lowest fitness value compared to the others, (iii) using the position of the fittest seagull as a reference, the rest can adjust their initial positions. Seagulls often attack migrating birds over the sea while moving from one location to another, employing a natural spiral movement during their attacks. Then, these behaviors can be formulated about an objective function for optimization purposes. 

During migration, the algorithm mimics the movement of the group of seagulls as they navigate from one position to another. In this phase, a seagull must meet three conditions: (i) Collision avoidance: To prevent collisions with neighboring seagulls, an additional variable is $A$ utilized in the calculation of the new position for the search agent:

\begin{equation}
\vec{C_s} = A \times \vec{P_s}(x)
\end{equation}
where \(\vec{C_s}\) denotes the position of the search agent that does not collide with other search agents, \(\vec{P_s}\) indicates the current position of the search agent, \(x\) refers to the current iteration, and \(A\) represents the movement behavior of the search agent within the specified search space. (ii) Movement toward the best neighbor's direction: After avoiding collisions with their neighbors, the search agents proceed in the direction of the best neighboring agent:
\begin{equation}
\vec{M_s} = B \times (\vec{P_{bs}}(x) - \vec{P_s}(x))
\end{equation}
where \(\vec{M_s}\) indicates the position of the search agent \(\vec{P_s}\) in relation to the best-fit search agent \(\vec{P_{bs}}\) (i.e., the fittest seagull). The behavior of \(B\) is randomized, which helps maintain an appropriate balance between exploration and exploitation. (iii) Stay close to the best search agent: Finally, the search agent can adjust its position in relation to the best search agent:
\begin{equation}
\vec{D_s} = |\vec{C_s} + \vec{M_s}|
\end{equation}
where \(\vec{D_s}\) denotes the distance between the search agent and the best-fit search agent (i.e., the best seagull with the lowest fitness value).

Exploitation aims to leverage the history and experiences gained during the search process. Seagulls have the ability to continuously adjust their angle of attack and speed while migrating. They regulate their altitude using their wings and body weight. When pursuing prey, they exhibit a spiral movement pattern in the air. This behavior in the x, y, and z planes is described as follows:
\begin{equation}
\begin{aligned}
x' &= r \times \cos(k)\\
y' &= r \times \sin(k)\\
z' &= r \times k\\
r &= u \times e^{kv}
\end{aligned}
\end{equation}
where \( r \) represents the radius of each turn of the spiral, \( k \) is a random number within the range \( [0 \leq k \leq 2\pi] \). \( u \) and \( v \) are constants that determine the spiral shape, and \( e \) is the base of the natural logarithm. The updated position of the search agent is calculated by Equation \ref{soa_pos}.
\begin{equation}
\vec{P_s}(x) = (\vec{D_s} \times x' \times y' \times z') + \vec{P_{bs}}(x)
\label{soa_pos}
\end{equation}
where \(\vec{P_s}(x)\) stores the best solution and updates the positions of the other search agents.

\citet{panagant2020seagull} introduced a surrogate-assisted metaheuristic approach for shape optimization, applying the SOA to optimize the structural shape of a vehicle bracket. The goal is to minimize structural mass while satisfying stress constraints. Finite element analysis (FEA) is used for function evaluations and is complemented by a Kriging model for estimation. Results indicate that SOA performs competitively, comparable to the whale optimization and salp swarm optimization algorithms, demonstrating strong potential for industrial component design. \citet{jia2019new} proposed three hybrid algorithms combining the SOA with thermal exchange optimization (TEO) for feature selection. The first algorithm employs a roulette wheel to alternate between SOA and TEO for location updates. The second method applies TEO after SOA iterations, while the third integrates TEO’s heat exchange formula into SOA’s attack mode to enhance exploitation capabilities. These hybrid algorithms demonstrate improved classification accuracy and efficient feature selection, achieving competitive results with reduced CPU time compared to existing hybrid optimization methods. \citet{dhiman2021mosoa} introduced the Multi-objective SOA (MOSOA). A dynamic archive is incorporated to store non-dominated Pareto optimal solutions, with a roulette wheel selection method to effectively select archived solutions by modeling seagull migration and attack behaviors. MOSOA is tested on twenty-four benchmark functions and compared against established metaheuristics. Additionally, it is applied to six constrained engineering design problems, demonstrating superior performance and high convergence of Pareto optimal solutions, making it well-suited for complex, real-world applications.
\subsection{Manta ray foraging optimization}

\begin{figure}[h]
\begin{center}  
\includegraphics[width=0.59\textwidth]{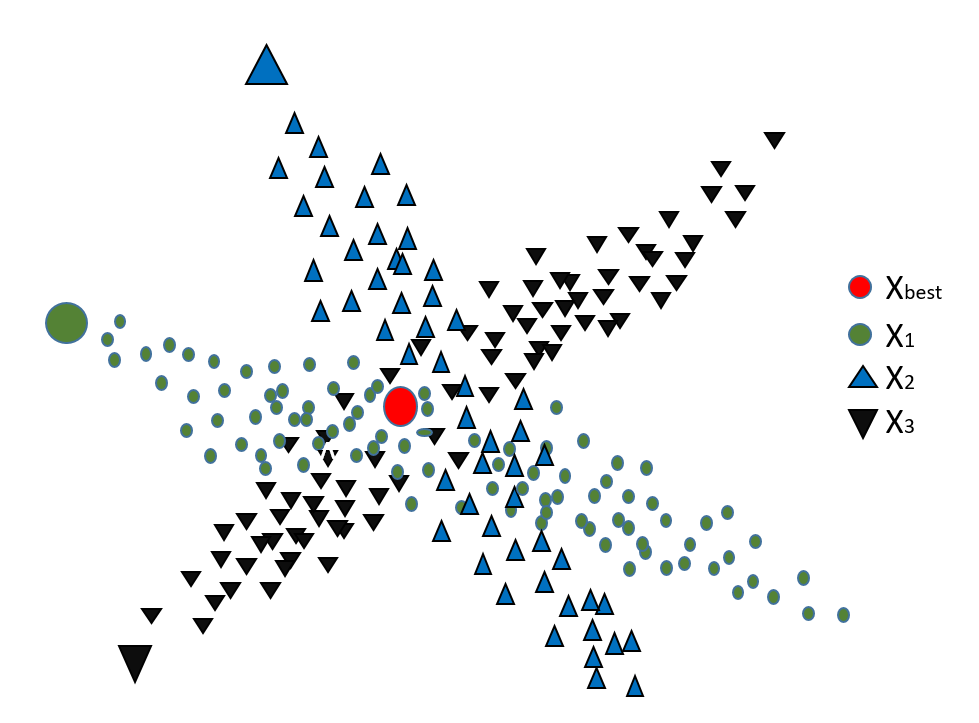}
\end{center} 
\caption{Somersault foraging behavior of three mantas in 2D} \label{manta}
\end{figure}

The Manta Ray Foraging Optimization (MRFO) algorithm is inspired by the foraging behavior of manta rays, focusing on both exploration and exploitation strategies through specific movements \citep{zhao2020manta}. See Figure \ref{manta} for the somersault foraging behavior of three mantas in 2 dimensions.

In the exploration phase, manta rays use a spiral movement to search for prey, modeled by the following update equation:
\[
\mathbf{X}_i^{t+1} = \mathbf{X}_i^t + A \cdot \left( \cos(\theta) \cdot (\mathbf{X}_i^t - \mathbf{X}_{\text{best}}^t) + \sin(\theta) \cdot (\mathbf{X}_i^t - \mathbf{X}_{\text{mean}}^t) \right),
\]
where \(A\) is a scaling factor, \(\theta\) is a random angle in the range \([0, 2\pi]\), \(\mathbf{X}_{\text{best}}^t\) is the position of the best solution found so far, and \(\mathbf{X}_{\text{mean}}^t\) is the mean position of the population. In the exploitation phase, manta rays move towards the best solution by adjusting their positions using the following equation:
\[
\mathbf{X}_i^{t+1} = \mathbf{X}_{\text{best}}^t + B \cdot (\mathbf{X}_i^t - \mathbf{X}_{\text{best}}^t),
\]
where \(B\) is a random factor influencing the movement toward the best solution.

Additionally, the manta ray may use a local search for the best solution:
\[
\mathbf{X}_i^{t+1} = \mathbf{X}_i^t + F \cdot (\mathbf{X}_{\text{best}}^t - \mathbf{X}_i^t),
\]
where \(F\) is a random factor that governs the intensity of the search.

Manta rays search for prey by adapting their movement towards the position of the prey. This is represented by:
\[
\mathbf{X}_i^{t+1} = \mathbf{X}_i^t + C \cdot (\mathbf{X}_{\text{prey}}^t - \mathbf{X}_i^t),
\]
where \(C\) is a scaling factor, and \(\mathbf{X}_{\text{prey}}^t\) is the position of the prey.

To avoid getting trapped in local optima, a diversification mechanism is applied, where each solution explores new regions of the search space:
\[
\mathbf{X}_i^{t+1} = \mathbf{X}_i^t + D \cdot (\mathbf{X}_{\text{worst}}^t - \mathbf{X}_i^t),
\]
where \(D\) is a scaling factor, and \(\mathbf{X}_{\text{worst}}^t\) is the position of the worst solution.

The MRFO algorithm iterates through exploration and exploitation phases until a termination criterion (e.g., maximum iterations or convergence threshold) is satisfied.

\citet{tang2021modified} presented a modified MRFO (m-MRFO) that enhances performance using an elite search pool, adaptive control parameters, and a distribution estimation strategy. Validated on 23 test functions and CEC2017, m-MRFO shows significant improvements and applicability to real-world engineering design problems. \citet{houssein2021improved} introduced the MRFO-OBL algorithm, an enhanced version of the MRFO (MRFO) algorithm that incorporates opposition-based learning (OBL) to improve population diversity and avoid local optima. MRFO-OBL addresses the segmentation of COVID-19 CT images using multilevel thresholding and is evaluated against six other metaheuristic algorithms, including the original MRFO. The results demonstrate that MRFO-OBL achieves superior quality, consistency, and robustness in segmentation, as measured by metrics like peak signal-to-noise ratio and structural similarity index, outperforming all compared algorithms. \citet{hassan2021improved} presented an innovative approach that combines MRFO with a Gradient-Based Optimizer (GBO) to tackle economic emission dispatch (EED) problems. This integration aims to enhance solution speed and reduce the likelihood of the original MRFO getting trapped in local optima. The MRFO–GBO addresses both single and multi-objective EED challenges while employing fuzzy set theory to identify optimal solutions in multi-objective scenarios. The algorithm is validated using CEC’17 test functions and applied to EED scenarios involving three electrical systems with different generator configurations. Results demonstrate that MRFO–GBO outperforms original MRFO and GBO, showcasing superior precision, robustness, and convergence characteristics in solving EED problems.

\subsection{Chimp optimization algorithm}

The Chimp Optimization Algorithm (ChOA) is inspired by the intelligent hunting and social cooperation behaviors of chimpanzees \citep{khishe2020chimp}. The algorithm models chimpanzee hunting strategies to balance exploration and exploitation during the optimization process. ChOA incorporates four main roles in chimpanzee groups: attackers, drivers, barriers, and chasers, each contributing to different search behaviors.

The initial population of chimpanzees is generated randomly. Chimpanzee hunting is simulated through the combined influence of the four groups (attackers, drivers, barriers, and chasers), and each group plays a different role in approaching the prey (solution). The position update rule is influenced by the best chimpanzee’s position:

\begin{equation}
X_i(t+1) = X_{\text{best}}(t) - A \cdot D
\end{equation}

where $X_{\text{best}}(t)$ is the position of the best chimpanzee at iteration $t$, $A$ is a coefficient that controls the direction and step size, and $D$ represents the distance between chimpanzee $i$ and the prey (best solution). The coefficient $A$ is calculated as follows:

\begin{equation}
A = 2 \cdot a \cdot r - a
\end{equation}

where $a$ decreases linearly from 2 to 0 over the course of iterations, balancing exploration and exploitation, and $r$ is a random number between 0 and 1. The distance between the chimpanzee and prey is calculated as:

\begin{equation}
D = |C \cdot X_{\text{best}}(t) - X_i(t)|
\end{equation}

where $C$ is another coefficient that controls the exploration phase and is calculated as:

\begin{equation}
C = 2 \cdot r
\end{equation}

Chimpanzees switch between exploration and exploitation based on the calculated coefficients and their role in the group. The different roles (attackers, drivers, barriers, and chasers) are represented mathematically to ensure a balance between the search mechanisms. The parameter $a$ decreases over time to transition the algorithm from exploration to exploitation, leading the chimpanzees toward better solutions as the iterations progress.

\begin{equation}
a(t) = 2 - \frac{t}{T}
\end{equation}

where: $t$ is the current iteration, and $T$ is the maximum number of iterations.

\citet{khishe2021weighted} proposed a weighted ChOA to address low convergence speed and local optima issues in large-scale numerical optimization. A position-weighted equation enhances convergence and avoids local optima, improving the balance between exploration and exploitation. Tested on 30 benchmark functions, IEEE competition benchmarks, and high-dimensional real-world problems, the proposed algorithm demonstrates superior performance in terms of speed and optimization accuracy. \citet{jia2021enhanced} introduced an Enhanced ChOA (EChOA) to improve solution accuracy. EChOA uses polynomial mutation for better population initialization, Spearman's rank correlation to compare chimps' social status, and a beetle antennae operator to improve exploration and avoid local optima. Tested on 12 classical benchmarks, 15 CEC2017 functions, and real-world engineering problems, EChOA outperforms ChOA and five other algorithms, demonstrating strong optimization capabilities and practical potential. \citet{du2022improved} presented an improved ChOa (IChOA) that integrates a somersault foraging strategy with adaptive weights to address 3D path planning challenges. The position vector updating equation is dynamically adjusted using a weighting factor derived from the original ChOA, while the somersault strategy helps prevent local optima and enhances early population diversity. Tested on CEC2019 functions and 3D path planning scenarios, IChOA demonstrates competitive performance compared to other methods. 
\subsection{Squirrel Search Algorithm}

\begin{figure}[h]
\begin{center}  
\includegraphics[width=0.65\textwidth]{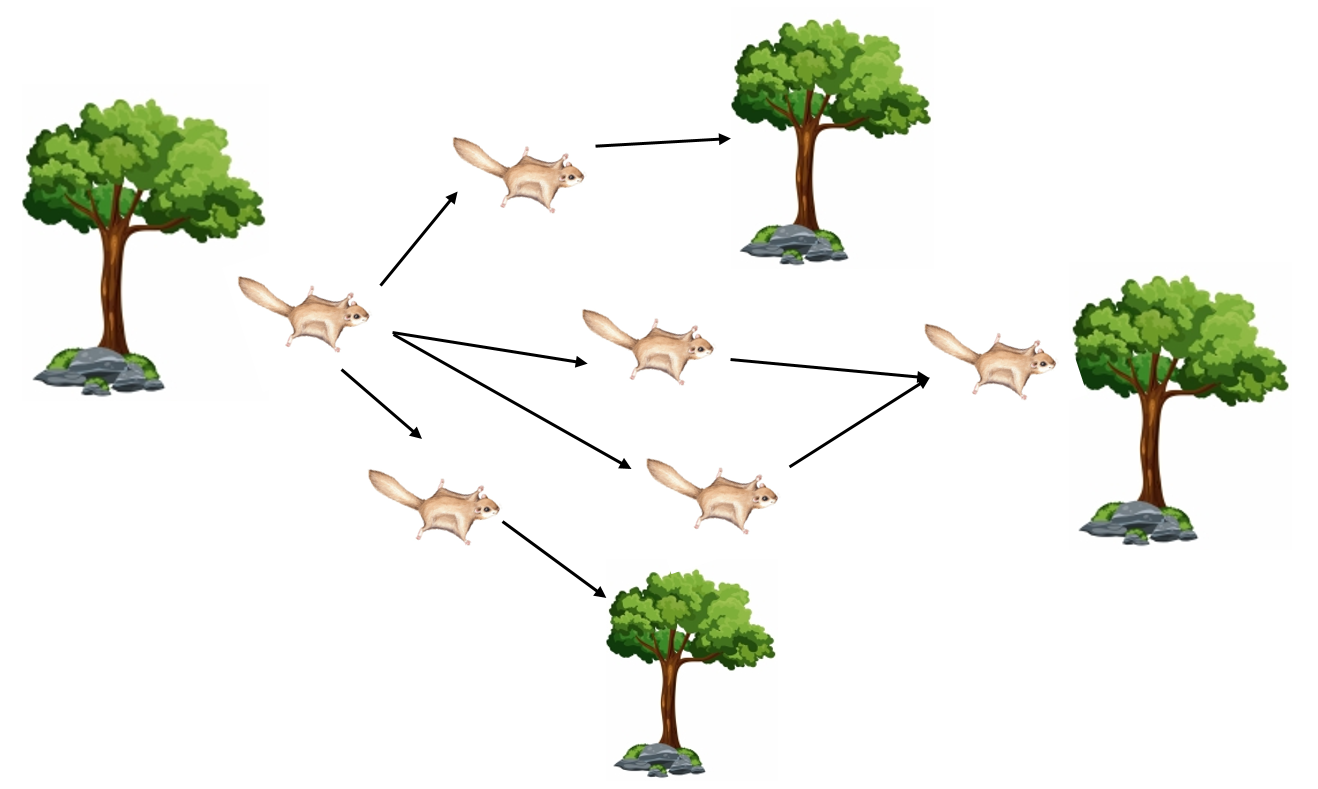}
\end{center} 
\caption{The fly behaviour of squirrels between trees.} \label{squirrel}
\end{figure}

The Squirrel Search Algorithm (SSA) mimics the foraging behavior of flying squirrels, which involves both exploration and exploitation \citep{jain2019novel}. The algorithm is characterized by random search movements and local search around discovered food sources. Figure \ref{squirrel} shows the fly behaviour of squirrels between trees.

In the exploration phase, squirrels search for food sources by moving randomly across the space. This is modeled as:
\[
\mathbf{X}_i^{t+1} = \mathbf{X}_i^t + \alpha \cdot \text{rand} \cdot (\mathbf{X}_i^t - \mathbf{X}_{\text{best}}^t),
\]
where \(\alpha\) is a scaling factor, \(\text{rand}\) is a random number between \([0, 1]\), and \(\mathbf{X}_{\text{best}}^t\) is the position of the best squirrel found so far.

In the exploitation phase, squirrels exploit the discovered food sources by performing a local search around the best food source found so far. The exploitation movement is given by:
\[
\mathbf{X}_i^{t+1} = \mathbf{X}_{\text{best}}^t + \beta \cdot (\mathbf{X}_i^t - \mathbf{X}_{\text{best}}^t),
\]
where \(\beta\) is a random factor that determines the intensity of the search around the best solution.

Alternatively, a squirrel may perform a random walk around the current solution:
\[
\mathbf{X}_i^{t+1} = \mathbf{X}_i^t + \gamma \cdot \text{rand} \cdot (\mathbf{X}_i^t - \mathbf{X}_{\text{mean}}^t),
\]
where \(\gamma\) is a scaling factor and \(\mathbf{X}_{\text{mean}}^t\) is the mean position of the population.

Squirrels evaluate food sources by their fitness. The position update is influenced by the fitness of the best food source:
\[
\mathbf{X}_i^{t+1} = \mathbf{X}_i^t + \delta \cdot (\mathbf{X}_i^t - \mathbf{X}_{\text{food}}^t),
\]
where \(\delta\) is a random scaling factor, and \(\mathbf{X}_{\text{food}}^t\) is the position of the food source that has the best fitness.

To prevent premature convergence, squirrels apply a diversification mechanism to explore other regions of the search space. This phase is modeled by:
\[
\mathbf{X}_i^{t+1} = \mathbf{X}_i^t + \epsilon \cdot (\mathbf{X}_{\text{worst}}^t - \mathbf{X}_i^t),
\]
where \(\epsilon\) is a diversification factor, and \(\mathbf{X}_{\text{worst}}^t\) is the position of the worst solution.

\citet{zheng2019improved}  introduces an improved SSA to enhance global convergence. ISSA incorporates several modifications: an adaptive predator presence probability to balance exploration and exploitation, a normal cloud model to capture the randomness in foraging, a successive position selection strategy to retain the best positions, and a dimensional search enhancement for improved local search. Tested on 32 benchmark functions, including unimodal, multimodal, and CEC 2014 functions, ISSA demonstrates competitive performance, outperforming the basic SSA and four other state-of-the-art algorithms. \citet{dhaini2021squirrel} applied the SSA to solve unconstrained and constrained portfolio optimization problems. Portfolio optimization seeks the best asset allocation, traditionally addressed by the Mean-Variance model (Markowitz) and its extensions, including the Sharpe model. Leveraging the success of nature-inspired algorithms, this study adapts SSA for both problem types, comparing it with various classical, hybrid, and multi-objective approaches. Results indicate that SSA excels in unconstrained optimization and performs competitively in constrained scenarios, achieving superior performance on different models and evaluation metrics. \citet{sakthivel2021combined} introduced a multi-objective SSA to address the combined economic and environmental power dispatch problem, an area gaining attention due to environmental concerns. The proposed SSA integrates Pareto dominance to produce non-dominated solutions, using an external elitist depository with crowding distance sorting to ensure diverse Pareto-optimal solutions. Tested on three complex systems, the algorithm demonstrates superior trade-offs between cost and emissions compared to other advanced heuristic methods.

\subsection{Henry gas solubility optimization}

Henry Gas Solubility Optimization (HGSO) is a metaheuristic inspired by the behavior of gas molecules in a liquid solution, based on Henry's Law \citep{hashim2019henry}. The algorithm simulates how gas particles interact and converge toward an optimal solution in the solution space. In HGSO, each molecule adjusts its position based on its concentration and the solubility coefficient \( k_H \) (Henry's constant), which impacts the movement intensity of each molecule.

The concentration \( C_i \) for each molecule is updated to reflect its solution quality:
\[
C_i^{(t+1)} = C_i^{(t)} + \alpha \left( C_{\text{best}} - C_i^{(t)} \right)
\]
where \( C_{\text{best}} \) is the concentration of the best solution found so far, and \( \alpha \) is a learning factor that controls the influence of \( C_{\text{best}} \) on \( C_i \).

The solubility of each molecule is affected by Henry's coefficient, which is updated using an evaporation function:
\[
k_H^{(t+1)} = k_H^{(t)} \cdot e^{-\beta \cdot t}
\]
where \( \beta \) is a decay parameter, and \( t \) is the current iteration. This coefficient modulates the search intensity, reducing as iterations progress to encourage convergence.

Each molecule's new position is determined by the concentration and solubility effects:
\[
\mathbf{x}_i^{(t+1)} = \mathbf{x}_i^{(t)} + k_H^{(t)} \cdot (C_{\text{best}} - C_i^{(t)}) \cdot \text{randn}(\mathbf{x}_i)
\]
where \( \text{randn}(\mathbf{x}_i) \) is a Gaussian-distributed random number applied to introduce controlled stochastic behavior.

The algorithm stops when a maximum number of iterations \( T \) is reached, or the change in the best solution is less than a threshold \( \epsilon \):
\[
\|\mathbf{x}_{\text{best}}^{(t+1)} - \mathbf{x}_{\text{best}}^{(t)}\| < \epsilon
\]

The HGSO algorithm employs a mechanism inspired by gas solubility in liquids, with concentration and solubility coefficients guiding the search process. By gradually reducing the exploration intensity, HGSO ensures effective convergence to the optimal solution.

\citet{neggaz2020efficient} presented a novel approach using the HGSO algorithm for FS, addressing challenges with large datasets prone to local optima. Tested on a variety of datasets with KNN and SVM classifiers, HGSO outperformed other metaheuristics like GOA and WOA. Statistical tests confirmed its effectiveness, achieving up to 100\% accuracy on datasets with over 11,000 features. \citet{yildiz2021novel} introduced the chaotic HGSO (CHGSO) algorithm, a metaheuristic that integrates chaotic maps into the original HGSO to enhance convergence for complex engineering optimization problems. Designed to be problem-independent, CHGSO was tested on various constrained optimization tasks, including welded and cantilever beam design, as well as automotive manufacturing and diaphragm spring design problems. Comparative results against established algorithms demonstrated CHGSO’s robustness and effectiveness in achieving optimal solutions across mechanical design and manufacturing challenges when paired with suitable chaotic maps. \citet{abd2021improved} presented a modified HGSO algorithm for optimal task scheduling in cloud computing. Integrating the Whale Optimization Algorithm (WOA) for local search and Comprehensive Opposition-Based Learning (COBL) for solution improvement, HGSWC aims to enhance task-to-resource mapping efficiency. Validated on 36 benchmark functions and tested on synthetic and real-world scheduling tasks, HGSWC outperformed conventional HGSO, WOA, and six other metaheuristic algorithms, achieving near-optimal solutions with minimal computational overhead.

\subsection{Archimedes optimization algorithm}

The Archimedes Optimization Algorithm (AOA) is a metaheuristic optimization algorithm inspired by Archimedes' principle, specifically buoyancy and density principles \citep{hashim2021archimedes}. The algorithm mimics the buoyant force acting on an object submerged in a fluid, which balances exploration (search for new solutions) and exploitation (refining existing solutions) by adjusting the density and volume of the solutions over time. The AOA algorithm generates initial candidate solutions randomly. The key component of the AOA is the buoyant force, which is calculated using the principle of buoyancy:

\begin{equation}
F_b = \rho \cdot V \cdot g
\end{equation}

where $F_b$ is the buoyant force, $\rho$ is the density of the fluid (related to the solution's quality), $V$ is the volume of the object (candidate solution), and $g$ is the gravitational acceleration constant. The positions of the candidate solutions are updated based on the calculated buoyant force and density:

\begin{equation}
X_i(t+1) = X_i(t) + F_b \cdot \left(1 - \frac{\rho}{\rho_{\text{max}}}\right)
\end{equation}

where $X_i(t+1)$ is the new position of solution $i$ at time $t+1$, $F_b$ is the buoyant force acting on the solution, and $\rho_{\text{max}}$ is the maximum allowable density, controlling the search's exploration and exploitation balance. The algorithm dynamically adjusts the density and volume of the solutions to balance exploration and exploitation. As the algorithm progresses, the density increases to focus on exploitation:

\begin{equation}
\rho = \rho_{\text{min}} + \left(\rho_{\text{max}} - \rho_{\text{min}}\right) \times \left(\frac{t}{T}\right)
\end{equation}

where $\rho_{\text{min}}$ and $\rho_{\text{max}}$ define the density range, $t$ is the current iteration number, and $T$ is the total number of iterations. The AOA includes an adaptive mechanism that adjusts the balance between exploration and exploitation over time. The following equation shows how the volume decreases as the algorithm converges:

\begin{equation}
V = V_{\text{max}} \times \left(1 - \frac{t}{T}\right)
\end{equation}

where $V_{\text{max}}$ is the maximum volume, the volume $V$ decreases as $t$ increases, encouraging exploitation in the later stages of the algorithm. 

\citet{yildiz2021comparision} examined the application of AOA in minimizing product development costs. It focuses on optimizing vehicle structures using size, shape, and topology optimization, demonstrating POA's superior search capability and computational efficiency.  \citet{akdag2022improved} proposed an Improved AOA (IAOA) to solve the Optimal Power Flow (OPF) problem. IAOA enhances population diversity and balances exploitation and exploration to prevent premature convergence. Tested on IEEE and South Marmara systems, the obtained simulation results and comparisons with different techniques show that the IAOA provides robustness in minimizing fuel emissions. \citet{desuky2021eaoa} introduced an Enhanced AOA (EAOA) for feature selection, improving the exploration-exploitation balance in the original AOA by adding a step-length parameter. Tested on twenty-three benchmark functions and sixteen real-world datasets, EAOA demonstrates superior classification performance and optimization results compared to AOA and other well-known algorithms. The results from sixteen real-world datasets confirm that the reduced feature subsets selected by the EAOA significantly enhance classification performance compared to other feature selection methods.
\subsection{Tunicate Swarm Algorithm}

The Tunicate Swarm Algorithm (TSA) is inspired by the collective behavior of tunicates, specifically their social interactions and foraging strategies \cite{kaur2020tunicate}. Tunicates, also known as sea squirts, exhibit unique behaviors that allow them to effectively find food and adapt to their environment. The TSA captures these behaviors to create an efficient search mechanism for solving complex optimization problems. Tunicates possess the ability to locate food sources in the ocean, yet there is no information available about the food source within the specified search area. This paper utilizes two behaviors observed in tunicates to locate optimal food sources: jet propulsion and swarm intelligence. 

To create a mathematical model for the jet propulsion behavior, the tunicate must meet three criteria: avoiding conflicts between search agents, moving towards the position of the most effective search agent, and staying close to that best search agent. In contrast, the swarm behavior facilitates the updating of other search agents' positions in relation to the optimal solution. To avoid conflicts between search agents that are other tunicates in the swarm, a vector \( \vec{A} \) is utilized to calculate the new positions of the search agents as shown in Formula \ref{tsa_avoid_a}.

\begin{equation}
\vec{A} = \vec{G} + \vec{M}
\label{tsa_avoid_a}
\end{equation}

\( \vec{G} \) represents the gravitational force, \( \vec{M} \) signifies the social forces acting between the search agents. The calculation for vector \( \vec{M} \) is expressed as:

\begin{equation}
\vec{M} = \left\lfloor P_{min} + c_1 \cdot (P_{max} - P_{min}) \right\rfloor
\label{tsa_avoid_m}
\end{equation}

\( P_{min} \) and \( P_{max} \) denote the initial and subordinate speeds that facilitate social interaction. In \cite{kaur2020tunicate}, \( P_{min} \) and \( P_{max} \) are set to 1 and 4, respectively. 
After conflicts have been resolved, the search agents then move toward the direction of their best neighbor. The calculation is represented by:

\begin{equation}
 \vec{PD} = \left| \vec{FS} - r_{and} \cdot \vec{P}_p(x) \right|
\end{equation}

where $\vec{PD}$ is the distance between the food source and the search agent (the tunicate), \( x \) indicates the current iteration, $\vec{FS}$ is the position of the food source (the optimum), $\vec{P}_p(x)$ signifies the position of the tunicate, and \( r_{and} \) is a random number within a specified range. Then, the search agent positions itself in relation to the best search agent (food source). The formula for this positioning is:

\begin{equation}
\vec{P}_p(x') = 
\begin{cases} 
\vec{F} \, S + \vec{A} \cdot \vec{P}_D, & \text{if } \text{rand} \geq 0.5 \\ 
\vec{F} \, S - \vec{A} \cdot \vec{P}_D, & \text{if } \text{rand} < 0.5 
\end{cases}
\end{equation}

where \( \vec{P}_p(x') \) is the updated position of the tunicate for the position of the food source \( \vec{FS} \). 
To mathematically simulate the swarm behavior of tunicates, the first two optimal solutions are recorded, enabling the update of the positions of other search agents based on the locations of the best agents. The following formula describes this swarm behavior:

\begin{equation}
\vec{P}_p(x + 1) =  \frac{\vec{P}_p(x) + \vec{P}_p(x + 1)}{2 + c_1}
\end{equation}

Search agents update their positions as \( \vec{P}_p(x') \), according to the best agents. The final position will be randomly located within a cylindrical or cone-shaped area defined by the position of the tunicate.


\citet{houssein2021improved} presented a TSA enhanced with a Local Escaping Operator (LEO) to address the limitations of the original TSA. The LEO strategy prevents search stagnation and enhances the convergence rate and local search efficiency of swarm agents. The effectiveness of the TSA-LEO was validated using the CEC'2017 test suite and compared against seven other metaheuristic algorithms. Results demonstrated that LEO significantly improves the solution quality and convergence speed of TSA. \citet{rizk2021enhanced} presented the Enhanced TSA (ETSA), an improvement on the TSA that enhances exploration and exploitation capabilities. ETSA was evaluated using 20 benchmark functions, including unimodal and multimodal tests, and compared with other algorithms. Statistical analyses confirmed its robustness and effectiveness, with the ETSA exhibiting resilience in high-dimensional scenarios and generally requiring less CPU time than competing methods. Finally, ETSA's applicability was demonstrated in the Economic Dispatch Problem, showcasing its effectiveness in real-world optimization tasks. \citet{gharehchopogh2022improved} aimed to enhance TSA's performance by incorporating mutating operators, specifically the Lévy, Cauchy, and Gaussian mutation operators, to address global optimization problems. The authors introduced a new algorithm, which leverages these operators, each contributing differently to the optimization process at various stages. The algorithm was tested on benchmark functions, including unimodal and multimodal groups, as well as six large-scale engineering problems. Experimental results demonstrate that the QLGCTSA algorithm outperforms competing optimization algorithms, showcasing its effectiveness in solving complex optimization tasks.

\subsection{Honey Badger Algorithm}

\begin{figure}[h]
\begin{center}  
\includegraphics[width=0.55\textwidth]{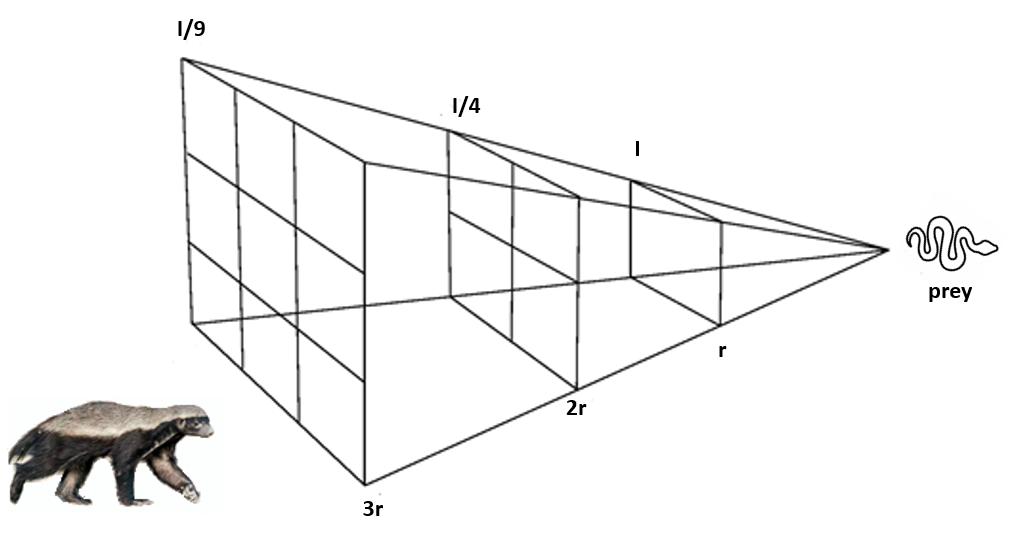}
\end{center} 
\caption{The inverse square law technique.} \label{honey-badge}
\end{figure}

The Honey Badger Algorithm (HBA) is inspired BY the foraging behavior and fearless nature of honey badgers \citep{hashim2022honey}. It is designed to solve complex optimization problems by utilizing a population of agents that explore the search space, balancing exploration and exploitation. The algorithm incorporates strategies such as local search, random movement, and hierarchical structure, allowing it to escape local optima and efficiently converge toward global solutions. HBA has been applied in various fields, including engineering, finance, and machine learning, demonstrating strong performance compared to other optimization algorithms. Figure \ref{honey-badge} shows the inverse square law technique used by honey badgers.

In the exploration phase, honey badgers perform random search movements to explore the search space. The position update equation is as follows:
\[
\mathbf{X}_i^{t+1} = \mathbf{X}_i^t + \alpha \cdot \text{rand} \cdot (\mathbf{X}_i^t - \mathbf{X}_{\text{best}}^t),
\]
where \(\alpha\) is a scaling factor, \(\text{rand}\) is a random number between \([0, 1]\), and \(\mathbf{X}_{\text{best}}^t\) is the best solution found so far.

In the exploitation phase, honey badgers exploit the best solution by updating their positions using a local search mechanism:
\[
\mathbf{X}_i^{t+1} = \mathbf{X}_{\text{best}}^t + \beta \cdot (\mathbf{X}_i^t - \mathbf{X}_{\text{best}}^t),
\]
where \(\beta\) is a random factor determining the movement towards the best solution.

Alternatively, if the honey badger is near a food source, it performs a more aggressive search:
\[
\mathbf{X}_i^{t+1} = \mathbf{X}_i^t + \gamma \cdot (\mathbf{X}_{\text{best}}^t - \mathbf{X}_i^t),
\]
where \(\gamma\) is a factor controlling the intensity of the aggressive search.

To simulate predator avoidance, honey badgers apply a diversification mechanism to avoid local optima and explore other regions of the search space. This phase is modeled as:
\[
\mathbf{X}_i^{t+1} = \mathbf{X}_i^t + \delta \cdot (\mathbf{X}_{\text{worst}}^t - \mathbf{X}_i^t),
\]
where \(\delta\) is a scaling factor, and \(\mathbf{X}_{\text{worst}}^t\) is the position of the worst solution found so far.

Honey badgers search for food by moving towards food sources. The update for food source attraction is:
\[
\mathbf{X}_i^{t+1} = \mathbf{X}_i^t + \epsilon \cdot (\mathbf{X}_{\text{food}}^t - \mathbf{X}_i^t),
\]
where \(\epsilon\) is a scaling factor, and \(\mathbf{X}_{\text{food}}^t\) is the position of the food source.

\citet{dli2022} focused on enhancing convergence in photovoltaic parameter estimation using two improved versions of the HBA. The first variant incorporates a Gauss/Mouse map-based chaotic approach to refine exploration and exploitation, while the second hybridizes opposition-based learning to scan the search space efficiently. Evaluated on CEC2017 and CEC2019 datasets, these algorithms demonstrate strong performance in optimizing parameters for single-diode, double-diode, and various photovoltaic models, including poly-crystalline and mono-crystalline types. \citet{fathy2023efficient} presented an energy management scheme for microgrids (MG) that utilizes the HBA to optimize the scheduling of generation units, including photovoltaic (PV) systems, wind turbines (WT), microturbines (MT), fuel cells (FC), and battery storage. The HBA effectively balances exploration and exploitation, avoiding local optima in complex optimization problems. Three operational scenarios are analyzed: normal PV and WT generation, WT at rated power, and both at maximum limits. The study focuses on two objectives-reducing operating costs and minimizing pollutant emissions while comparing HBA's performance against various optimization algorithms. Results demonstrate HBA's superior robustness and effectiveness across all tested conditions, making it a strong candidate for enhancing microgrid operations.
\subsection{Mayfly optimization algorithm}

The Mayfly Optimization Algorithm (MOA) is inspired BY the swarming and mating behavior of mayflies \citep{ZERVOUDAKIS2020106559}. MOA simulates the dynamics between male and female mayflies to explore the solution space effectively. The population consists of male and female mayflies. The positions of the mayflies represent possible solutions, which evolve over time due to attraction, mating, and movement rules. The algorithm iterates until it converges to an optimal solution or reaches a maximum number of iterations.

Male mayflies update their velocities based on attraction to other males and a gravitational pull toward the fittest mayfly in the population. The velocity of a male mayfly \( \mathbf{M}^i \) is updated as:
\[
\mathbf{V}_M^i(t+1) = w \cdot \mathbf{V}_M^i(t) + r_1 \cdot \alpha \cdot \left(\mathbf{M}^{\text{best}} - \mathbf{M}^i(t)\right) + r_2 \cdot \beta \cdot \left(\mathbf{M}^j - \mathbf{M}^i(t)\right)
\]
where:
- \(w\) is the inertia weight,
- \(r_1\) and \(r_2\) are random numbers uniformly distributed in \([0, 1]\),
- \(\alpha\) and \(\beta\) are attraction coefficients,
- \(\mathbf{M}^{\text{best}}\) is the position of the fittest male mayfly,
- \(\mathbf{M}^j\) is the position of a neighboring male mayfly.

The new position of the male mayfly is then calculated as:
\[
\mathbf{M}^i(t+1) = \mathbf{M}^i(t) + \mathbf{V}_M^i(t+1)
\]

Female mayflies are attracted to their corresponding male partners. The velocity of a female mayfly \( \mathbf{F}^i \) is updated as:
\[
\mathbf{V}_F^i(t+1) = w \cdot \mathbf{V}_F^i(t) + r_3 \cdot \gamma \cdot \left(\mathbf{M}^i - \mathbf{F}^i(t)\right)
\]
where:
- \(w\) is the inertia weight,
- \(r_3\) is a random number uniformly distributed in \([0, 1]\),
- \(\gamma\) is an attraction coefficient for females toward males,
- \(\mathbf{M}^i\) is the position of the corresponding male mayfly.

The new position of the female mayfly is then calculated as:
\[
\mathbf{F}^i(t+1) = \mathbf{F}^i(t) + \mathbf{V}_F^i(t+1)
\]

When the distance between a male and a female mayfly becomes small (i.e., when they are close in the search space), mating occurs, and offspring are generated. The offspring inherits characteristics from both parents, with the initial position of the offspring given by:
\[
\mathbf{O}^i = \delta \cdot \mathbf{M}^i + (1 - \delta) \cdot \mathbf{F}^i
\]
where \( \delta \in [0, 1] \) is a weighting factor that controls the contribution of each parent to the offspring's position.

The algorithm iterates until it meets a termination condition, which could be a maximum number of iterations or a satisfactory fitness level for the solutions.

\citet{gao2020improved} combined the MOA with PSO and Differential Evolution (DE), with improved velocity updates based on Cartesian distances, enhancing individuals’ movement toward each other. Simulations show that this revised MO version outperforms the original, offering better optimization and convergence. \citet{shaheen2021precise} introduced the Chaotic MOA (CMOA) to accurately model proton exchange membrane fuel cells (PEMFCs). By optimizing seven design variables absent in manufacturer data, CMOA minimizes the total squared error between laboratory-measured and simulation-derived voltages, addressing PEMFC non-linear I-V characteristics. Integrating chaotic mapping with MOA enhances solution quality. The model, tested across different PEMFC types and conditions (temperature, pressure), shows that CMOA achieves precise simulations, verified against other optimization methods for robust and reliable PEMFC modeling. \citet{nagarajan2022combined} presented an improved MOA (IMA) using Levy flight to address the combined economic emission dispatch (CEED) problem in microgrids, aimed at optimizing generation cost and minimizing emissions. The study focuses on an islanded microgrid setup with thermal, solar, and wind power sources, testing over a 24-hour period with varying demand. IMA achieves significant reductions in both cost and emissions across four scenarios, outperforming the original Mayfly algorithm and other methods, highlighting its effectiveness in optimizing CEED for grid-connected microgrids.

\subsection{African vultures optimization}

\begin{figure}[h]
\begin{center}  
\includegraphics[width=0.59\textwidth]{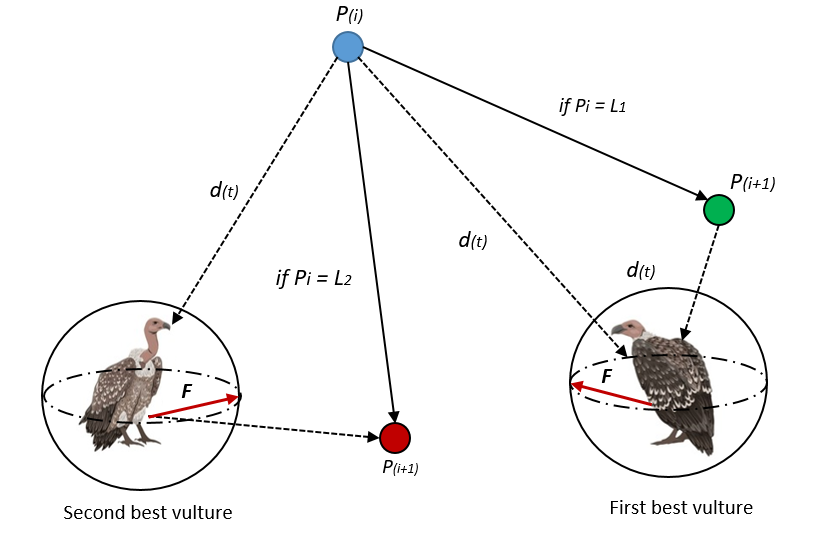}
\end{center} 
\caption{Overall vectors of African vultures in the case of competition for food.} \label{African}
\end{figure}

The African Vulture Optimization Algorithm (AVOA) is inspired by the scavenging behavior of vultures \cite{abdollahzadeh2021african}. The algorithm mimics how African vultures search for carcasses by exploring large areas and converging when a food source is detected. In AVOA, vultures are represented as agents that search for optimal solutions in a problem space, with each agent adjusting its position based on exploration (searching for new areas) and exploitation (focusing on known promising regions). AVOA's balance between exploration and exploitation makes it suitable for solving optimization problems efficiently across various domains, such as engineering design, machine learning, and scheduling (See Figure \ref{African} for the overall vectors of African vultures in the case of competition for food). 

The position of each vulture is updated in each iteration based on the current location of the prey (best solution). The equation for position update is:

\begin{equation}
X_i(t+1) = X_{\text{prey}}(t) + A \cdot D
\end{equation}

where $X_i(t+1)$ is the updated position of vulture $i$ at time step $t+1$, $X_{\text{prey}}(t)$ is the current position of the prey, and  $A$ and $D$ are adaptive parameters that control the influence of the prey’s position. The distance between each vulture and the prey is calculated to guide the movement. This distance is expressed as:

\begin{equation}
D = |C \cdot X_{\text{prey}}(t) - X_i(t)|
\end{equation}

$D$ represents the distance between vulture $i$ and the prey, and $C$ is a coefficient that scales the distance based on environmental factors.

In some algorithm variations, vultures can move in a spiral around the prey to simulate a more complex search. The spiral motion is described by:

\begin{equation}
X_i(t+1) = D \cdot e^{b \cdot l} \cdot \cos(2\pi l) + X_{\text{prey}}(t)
\end{equation}

$b$ controls the width of the spiral, $l$ is a random variable controlling the angle of the spiral, and $e$ is the base of the natural logarithm, indicating the exponential nature of the spiral. To balance exploration and exploitation, the algorithm employs two adaptive coefficients, $A$ and $C$, which change over time. These coefficients are defined as:

\begin{equation}
A = 2 \cdot a \cdot \text{rand}(0,1) - a
\end{equation}

\begin{equation}
C = 2 \cdot \text{rand}(0,1)
\end{equation}

where $a$ is a variable that decreases over time, promoting exploitation as the algorithm converges.

\citet{askr2023many} presented a novel many-objective AVOA (MaAVOA), incorporating a new social leader vulture in the selection process and an alternative pool-based environmental selection mechanism. Through experiments on DTLZ functions \cite{tanabe2017note} and real-world problems, MaAVOA demonstrates superior convergence, diversity, and statistical relevance performance compared to existing algorithms, making it a promising solution for complex engineering problems.  \citet{alanazi2022optimal} studied the performance of photovoltaic (PV) systems that are heavily influenced by weather conditions like irradiance and temperature, with partial shade conditions (PSC) causing issues such as hot spots and power loss. They introduced the AVOA to optimize PV array reconfiguration under PSC to maximize power generation. Comparative studies across five shading patterns show that AVOA outperforms methods in power enhancement and performance ratio. Chaotic mapping is recommended to fine-tune AVOA’s parameters for improved results. \citet{fan2021improved} developed a new metaheuristic algorithm (TAVOA) that enhances the AVOA by using tent chaotic mapping for population initialization and a time-varying mechanism to balance exploration and exploitation. Tested on benchmark functions and real-world engineering problems, TAVOA significantly outperforms AVOA and other state-of-the-art algorithms in multiple cases, demonstrating its improved optimization capabilities.

\subsection{Golden jackal optimization}

\begin{figure}[h]
\begin{center}  
\includegraphics[width=0.6\textwidth]{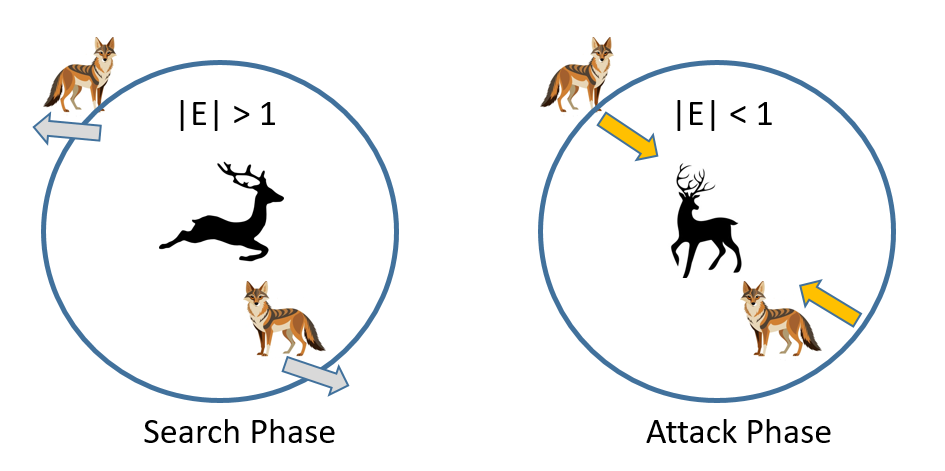}
\end{center} 
\caption{Searching and attacking phases of the golden jackal.} \label{golden}
\end{figure}

The Golden Jackal Optimization (GJO) algorithm is a recent population-based metaheuristic inspired by the hunting behavior and social hierarchy of golden jackals \citep{chopra2022golden}. Similar to other nature-inspired algorithms, GJO models the balance between exploration and exploitation during the search process, mimicking the way golden jackals collaboratively hunt prey and share resources. Se Figure \ref{golden} for the phases of searching and attacking of the golden jackal.

The GJO algorithm leverages several key mechanisms to optimize complex problems. In the exploration phase, jackals search the solution space by mimicking random and collective movement patterns in search of prey, aiming to avoid local optima. Once potential solutions are located, jackals focus on refining and exploiting the most promising areas of the solution space, akin to converging toward a prey's location. The hierarchical structure of the golden jackal pack influences the decision-making process, with higher-ranked jackals guiding the search direction based on successful previous experiences.

GJO has demonstrated effectiveness in solving various optimization problems, such as feature selection, engineering design, and multi-objective optimization. Its performance is often compared to other metaheuristic algorithms like PSO and Genetic Algorithms (GA), showing competitive results in balancing convergence speed and solution accuracy.

The mathematical model of GJO can be expressed as follows:

\[
\mathbf{X}_{t+1} = \mathbf{X}_t + \alpha \cdot \left( \text{Rand} \cdot (\mathbf{L}_{\text{best}} - \mathbf{X}_t) + (1-\text{Rand}) \cdot (\mathbf{G}_{\text{best}} - \mathbf{X}_t) \right)
\]

where: $\mathbf{X}_t$ is the current position of the jackal at iteration $t$, $\alpha$ is a control parameter influencing the movement step size, $\text{Rand}$ is a random number between 0 and 1, and $\mathbf{L}_{\text{best}}$ and $\mathbf{G}_{\text{best}}$ represent the local and global best solutions, respectively.

\citet{yuan2022hybrid} proposed a hybrid GJO algorithm (LSGJO) by integrating the Gold-SA and dynamic lens-imaging learning. New update rules and scaling factors improve population diversity, avoiding local optima. LSGJO outperforms other algorithms in both benchmark functions and real-world design problems, with faster, more accurate convergence. Experimental results demonstrate that LSGJO outperforms 11 cutting-edge optimization algorithms, achieving faster and more precise convergence. The algorithm significantly enhances both global and local search capabilities and excels in solving complex constrained problems.  \citet{rezaie2022model}  employed a PSO-based GJO method to minimize the sum of squared errors (SSE) between the measured and simulated output voltages of the PEMFC stack. The proposed approach is validated on two cases and compared with various recent optimizers, showing that ICSO delivers superior performance in estimating the optimal PEMFC model. GJO struggles with weak exploitation, local optima, and balancing exploration and exploitation. To address this, \citet{mohapatra2023fast}  introduced the fast random opposition-based learning GJO (FROBL-GJO), enhancing precision and convergence speed using opposition-based learning techniques. Tested on CEC benchmarks and real-world problems, FROBL-GJO outperforms other methods, proving its effectiveness in global optimization and engineering design.

\subsection{Dung beetle optimizer}

The Dung Beetle Optimizer (DBO) is inspired by the rolling behavior of dung beetles, which involves finding and transporting dung to create nests \citep{xue2023dung}. Dung beetles show intelligent foraging strategies by navigating and rolling dung balls in optimal directions. The DBO algorithm mimics this behavior, balancing exploration and exploitation through a combination of directed movements and random perturbations. The movement update of the solutions is given by:

\begin{equation}
X_i(t+1) = X_{\text{best}}(t) + A \cdot D
\end{equation}

Where $X_{\text{best}}(t)$ is the position of the best solution at iteration $t$, $A$ is a coefficient controlling the direction of movement, and $D$ represents the distance between the dung beetle and the best solution. The movement coefficient $A$ is updated to balance exploration and exploitation:

\begin{equation}
A = 2 \cdot a \cdot r - a
\end{equation}

Where: $a$ is a parameter that decreases over time, encouraging convergence, and $r$ is a random number between 0 and 1.

The distance $D$ is calculated as:

\begin{equation}
D = |C \cdot X_{\text{best}}(t) - X_i(t)|
\end{equation}

Where $C$ is another coefficient controlling the distance and is calculated as:

\begin{equation}
C = 2 \cdot r
\end{equation}

DBO uses adaptive control of the parameters $A$ and $C$ to balance exploration and exploitation. This allows the algorithm to explore the search space initially and focus on exploitation in later stages. As iterations progress, the algorithm encourages convergence towards the best solutions through decreasing values of $a$:

\begin{equation}
a(t) = 2 - \frac{t}{T}
\end{equation}

where $t$ is the current iteration, and $T$ is the maximum number of iterations.

\citet{duan2023air} presented a combined model for predicting the Air Quality Index (AQI) using real data from four cities, employing an ARIMA model for the linear component and a CNN-LSTM model for the non-linear component, with hyperparameters optimized using the DBO. The proposed model outperformed nine widely used models. \citet{shen2023multi} introduced a multi-strategy enhanced DBO (MDBO) to improve the original DBO by addressing its limitations in global search capability and local optima trapping. The MDBO employs a dynamic Beta distribution for reflection solutions, a Levy distribution to manage out-of-bounds particles, and two cross operators to enhance the updating process of the algorithm, resulting in improved convergence and a better balance between exploration and exploitation. \citet{jaiswal2023dung} presented the DBO for solving the optimal power flow (OPF) problem with the integration of solar and wind energy sources. Given the stochastic nature of these energy sources, the DBO takes into account their uncertainty by utilizing Log-normal and Weibull probability density functions to estimate solar and wind power generation. The effectiveness of the DBO algorithm is demonstrated through its implementation on both standard and modified IEEE 30-bus systems in MATLAB, with extensive comparative analyses against various optimization methods showcasing its reliability and effectiveness in addressing complex power system challenges. 
\subsection{Coati Optimization Algorithm}

\begin{figure}[h]
\begin{center}  
\includegraphics[width=0.69\textwidth]{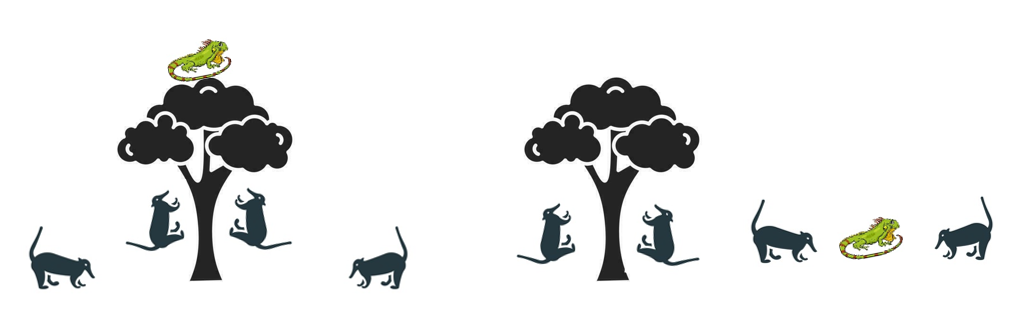}
\end{center} 
\caption{Attack of the coatis’ population to an iguana on the tree and hunting fallen iguana on the ground by the other half.} \label{caoti}
\end{figure}

The Coati Optimization Algorithm (COA) is inspired by the social foraging and movement patterns of coatis, which are small omnivorous mammals known for their cooperative behavior and intelligent problem-solving skills \citep{dehghani2023coati}. COA incorporates two key phases: exploration and exploitation, which are governed by the social hierarchy and foraging habits of coatis. During the exploration phase, coatis moves to discover new areas of the search space. See Figure \ref{caoti} for the attack of the coatis on an iguana on the tree and hunting fallen iguana on the ground by the other half.

Their movement is modeled as:

\begin{equation}
X_i(t+1) = X_i(t) + \beta \cdot (X_{\text{leader}}(t) - X_i(t)) + \alpha \cdot \text{randn}(0,1)
\end{equation}

Where  $X_{\text{leader}}(t)$ is the position of the leading coati at iteration $t$, $\beta$ controls the influence of the leader, $\alpha$ is a coefficient that adds random perturbations to encourage exploration, and $\text{randn}(0,1)$ is a normally distributed random number. In the exploitation phase, coatis focuses on fine-tuning their positions around the best-known solutions. The position of each coati is updated as:

\begin{equation}
X_i(t+1) = X_i(t) + \gamma \cdot (X_{\text{best}}(t) - X_i(t))
\end{equation}

Where $X_{\text{best}}(t)$ is the best solution found so far, and $\gamma$ is a parameter that controls the step size towards the best solution. COA uses an adaptive behavior mechanism where parameters such as $\beta$, $\alpha$, and $\gamma$ are adjusted over time to balance exploration and exploitation. The adaptive mechanism is formulated as:

\begin{equation}
\gamma(t) = \gamma_{\text{min}} + (\gamma_{\text{max}} - \gamma_{\text{min}}) \cdot \frac{t}{T}
\end{equation}

Where $\gamma_{\text{min}}$ and $\gamma_{\text{max}}$ define the range for the exploitation parameter, $t$ is the current iteration, and $T$ is the total number of iterations.

\citet{hashim2023efficient} introduced the Dynamic COA (DCOA) as a feature selection technique that iteratively introduces different features during the optimization process. DCOA enhances exploration and exploitation capabilities through dynamic opposing candidate solutions and requires no preparatory parameter tuning. Evaluated on the CEC’22 test suite and nine medical datasets, DCOA demonstrated superior performance compared to seven well-known metaheuristic algorithms, achieving an overall accuracy of 89.7\%, a feature selection rate of 24\%, sensitivity of 93.35\%, specificity of 96.81\%, and precision of 93.90\%, as confirmed by statistical tests. \citet{bacs2023enhanced} introduced an enhanced COA (ECOA) that incorporates two modifications to maintain population diversity during searches. Evaluated across various test groups, ECOA outperformed COA on twenty-three classic CEC functions, CEC-2017, and CEC-2020 functions in multiple dimensions (5, 10, and 30). It also excelled in Big Data Optimization Problems (BOP) across different cycles (300, 500, and 1000). Statistical tests confirmed ECOA's superior performance compared to COA and seven other recently proposed algorithms, making it a strong alternative for continuous optimization problems.  \citet{hashim2023efficient} introduced a modified COA (mCoatiOA), which enhances the original algorithm by incorporating adaptive s-best mutation, directional mutation, and search direction control toward the global best. Tested against various optimization algorithms on the CEC'20 test suite and fifteen benchmark datasets from the UCI repository, mCoatiOA outperformed competitors, achieving the best results on 75\% of the datasets and demonstrating significant improvements in average fitness and standard deviation values.

\subsection{Chaos Game Optimization}

Chaos Game Optimization (CGO) is a metaheuristic algorithm inspired by the concept of the chaos game, where a point iteratively moves closer to randomly chosen vertices of a fractal structure, generating a pattern that covers a fractal attractor \citep{talatahari2021chaos}.

CGO begins by initializing a population of points $\mathbf{x}_i$ randomly within the feasible search space:
\[
\mathbf{x}_i = \mathbf{x}_{\text{min}} + \text{rand}(\mathbf{x}_{\text{max}} - \mathbf{x}_{\text{min}})
\]
where $\mathbf{x}_{\text{min}}$ and $\mathbf{x}_{\text{max}}$ define the lower and upper bounds of the search space, respectively, and \( \text{rand} \) is a function that generates a random number within \([0,1]\).

The position update rule in CGO relies on a target point $\mathbf{g}$ and the use of a chaotic map. Let $\mathbf{g}$ represent a randomly selected point from the current population or the best-so-far solution. The new position of each solution $\mathbf{x}_i$ is updated as follows:
\[
\mathbf{x}_i^{(t+1)} = \mathbf{x}_i^{(t)} + \beta (\mathbf{g} - \mathbf{x}_i^{(t)})
\]
where \( \beta \) is a scaling factor that determines the step size and can be tuned based on a chaotic map. A commonly used chaotic map is the logistic map:
\[
\beta_{t+1} = r \beta_t (1 - \beta_t)
\]
where \( r \) is a parameter (typically \( r = 4 \) for chaotic behavior) and \( \beta \in (0, 1) \).

The algorithm iterates until a termination criterion is met, such as a maximum number of iterations \( T \) or when the change in the best solution is smaller than a predefined threshold \( \epsilon \):
\[
\|\mathbf{x}_{\text{best}}^{(t+1)} - \mathbf{x}_{\text{best}}^{(t)}\| < \epsilon
\]

The CGO algorithm leverages chaotic behavior and fractal properties to explore the search space and converges toward optimal solutions. The chaotic map enhances diversity and aids in escaping local optima.

\cite{talatahari2020optimization} tested CGO algorithm on 34 benchmarked constrained mathematical problems and 15 engineering design problems. The results were compared with other standard, improved, and hybrid metaheuristic algorithms, using statistical measures such as minimum, mean, maximum, and standard deviation. The CGO algorithm demonstrated competitive performance, outperforming other metaheuristics in most cases. \citet{khodadadi2023multi} proposed multi-objective CGO (MOCGO) that stores Pareto-optimal solutions in a fixed-sized external archive and incorporates leader selection for multi-objective optimization. The algorithm is applied to eight real-world engineering design challenges with multiple objectives, using chaos theory and fractal models inherited from CGO. Performance is assessed through seventeen case studies, including CEC-09, ZDT, and DTLZ, and compared to six well-known multi-objective algorithms using four performance metrics. The results show that MOCGO outperforms existing methods, achieving excellent convergence and coverage of Pareto-optimal solutions. \citet{ramadan2021new} introduced the CGO algorithm for estimating the unknown parameters of the three-diode (TD) photovoltaic (PV) model, which is crucial for enhancing the accuracy of PV energy system simulations. The PV model is highly nonlinear, and the lack of complete parameter information in PV cell datasheets complicates its modeling. The proposed CGO-based method is used to estimate these parameters for real PV cells and modules, varying temperature and irradiation conditions. The simulation results are compared with experimental data to validate the model's accuracy. The CGO algorithm demonstrates the lowest Root Mean Square Error (RMSE), mean, and standard deviation, and provides the fastest implementation time compared to other existing techniques.

\subsection{Beluga whale optimization}

The Beluga Whale Optimization (BWO) algorithm is a nature-inspired metaheuristic that mimics the hunting and migratory behavior of beluga whales. BWO balances exploration and exploitation by modeling the unique spiral-shaped hunting path and collective migration of belugas in search of food. The algorithm adjusts its parameters over time to converge toward optimal solutions efficiently. Beluga whales are randomly positioned within the search space at the start of the algorithm. The key feature of the BWO is the spiral-shaped path used to simulate hunting behavior:

\begin{equation}
X_i(t+1) = X_{\text{best}}(t) + D \cdot e^{b \cdot l} \cdot \cos(2\pi l)
\end{equation}

where $X_{\text{best}}(t)$ is the position of the best solution at time $t$, $D$ is the distance between the whale and the prey, $b$ controls the tightness of the spiral, and $l$ is a random variable, dictating the spiral shape. Beluga whales also engage in collective migration, which influences the algorithm’s exploitation phase. Each whale adjusts its position based on both its current location and the best-known position in the group:

\begin{equation}
X_i(t+1) = X_{\text{best}}(t) + C \cdot (X_{\text{best}}(t) - X_i(t)) 
\end{equation}

Where $C$ is a coefficient that controls the attraction toward the best-known solution, $X_{\text{best}}(t)$ is the position of the best whale at time $t$, and $X_i(t)$ is the current position of whale $i$. The BWO algorithm uses adaptive parameters to balance exploration and exploitation. Over time, the spiral path becomes tighter, and the attraction toward the best solution increases:

\begin{equation}
b(t) = b_{\text{min}} + (b_{\text{max}} - b_{\text{min}}) \cdot \left(1 - \frac{t}{T}\right)
\end{equation}

where $b(t)$ decreases as the number of iterations increases, and $t$ is the current iteration, and $T$ is the total number of iterations.

\citet{li2024multi} presented a multi-objective hierarchical optimal planning model for distributed generation (DG) using an improved BWO (IBWO) algorithm. It considers DG output uncertainties and demand response, often neglected in past research. The IBWO algorithm effectively reduces annual comprehensive costs, voltage deviation, and power losses by 11.66\%, 40.55\%, and 38.61\%, respectively, enhancing power quality and economic efficiency. \citet{hussien2023novel} introduced a modified BWO (mBWO) algorithm, addressing limitations of the original BWO, such as premature convergence and imbalance between exploration and exploitation. mBWO incorporates elite evolution, randomization control, and a transition factor to enhance performance. It outperforms the original BWO and 10 other optimizers on 29 CEC2017 functions and eight engineering design problems, delivering superior results in constrained and unconstrained environments. \citet{houssein2023dynamic} enhances the BWO algorithm to address its lack of diversity and premature convergence. The improved BWO uses Opposition-Based Learning (OBL) and Dynamic Candidate Solutions (DCS) with the k-Nearest Neighbor (kNN) classifier. The enhanced OBWOD algorithm is tested on CEC’22 benchmarks and 10 medical datasets, outperforming seven algorithms with an overall classification accuracy of 85.17\%, demonstrating its competitive performance.
\subsection{Gazelle optimization algorithm}

The Gazelle Optimization Algorithm (GOA) is inspired by the behavior of gazelles in the wild, particularly their fast, adaptive movements and their ability to avoid predators by using sharp turns and rapid acceleration \citep{agushaka2023gazelle}. This algorithm mimics the dynamic strategies gazelles use to explore and exploit their environment efficiently, balancing exploration and exploitation by adjusting movement patterns throughout the optimization process. The position of each gazelle is updated based on the best-known positions in the population, using adaptive movement strategies inspired by gazelle dynamics. The position update equation is given by:

\begin{equation}
X_i(t+1) = X_{\text{best}}(t) + V_i(t) + \alpha \cdot (X_{\text{best}}(t) - X_i(t))
\end{equation}

Where: $X_{\text{best}}(t)$ is the best-known position at iteration $t$, $V_i(t)$ represents the velocity of the gazelle at iteration $t$, and $\alpha$ is a parameter controlling the intensity of the attraction towards the best-known position. The velocity of each gazelle is updated to ensure a balance between exploration and exploitation. The velocity update is influenced by the difference between the gazelle’s current position and the best-known position:

\begin{equation}
V_i(t+1) = \beta \cdot V_i(t) + \gamma \cdot (X_{\text{best}}(t) - X_i(t)) + \delta \cdot (X_{\text{rand}}(t) - X_i(t))
\end{equation}

Where:  $\beta$ controls the inertia of the velocity, $\gamma$ influences the attraction towards the best-known position, and $\delta$ controls the influence of a random solution $X_{\text{rand}}(t)$ to maintain diversity. The parameters $\alpha$, $\beta$, and $\gamma$ are adjusted dynamically throughout the optimization process to control the balance between exploration and exploitation. Typically, these parameters decrease over time to allow for initial exploration followed by exploitation:

\begin{equation}
\alpha(t) = \alpha_{\text{max}} - \frac{t}{T} \cdot (\alpha_{\text{max}} - \alpha_{\text{min}})
\end{equation}

\begin{equation}
\beta(t) = \beta_{\text{max}} - \frac{t}{T} \cdot (\beta_{\text{max}} - \beta_{\text{min}})
\end{equation}

Where: $t$ is the current iteration, and $T$ is the total number of iterations. As iterations progress, the gazelles are increasingly attracted towards the best-known solutions, with decreasing random exploration. This results in convergence towards optimal solutions while maintaining diversity:

\begin{equation}
X_{\text{best}}(t+1) = X_{\text{best}}(t) + \epsilon \cdot (X_{\text{best}}(t) - X_{\text{mean}}(t))
\end{equation}

Where: $X_{\text{mean}}(t)$ is the mean position of all gazelles at iteration $t$, and $\epsilon$ controls the influence of the mean position on the best solution.

\citet{khodadadi2023mountain} proposed the mountain (MGO) as a new metaheuristic algorithm for optimizing truss structures in structural engineering. Inspired by gazelle social behavior, MGO aims to handle complex, constrained design problems characterized by multiple local optima and non-convex search spaces, offering optimal, lightweight design solutions compared to traditional optimization methods. \citet{mehta2024new} applied the mountain MGO, inspired by gazelle social behaviors, and a neural network to optimize vehicle components and other mechanical systems. By hybridizing MGO with the Nelder–Mead algorithm (HMGO-NM), it tackles automotive, manufacturing, construction, and mechanical engineering tasks. Comparative results indicate HMGO-NM's superiority, showing broad potential across industrial applications. \citet{abdel2024adaptive} presented the adaptive chaotic dynamic GOA (ACD-GOA), an advanced version of the GOA tailored for feature selection (FS). ACD-GOA enhances the search efficiency and convergence speed through dynamic opposition learning, adaptive inertia weights, and elite strategies. Evaluations on twelve CEC2022 functions and fourteen FS benchmarks show ACD-GOA's effectiveness, achieving classification accuracy between 0.78 and 1.00 with the K-NN classifier, and outperforming other metaheuristic algorithms.

\section{State-of-the-art Applications of Recent Metaheuristics}\label{section4}

This section outlines the recent applications of metaheuristic algorithms, primarily focusing on their use from 2019 to 2024. Metaheuristics have become vital in optimizing complex systems in various domains, including engineering, healthcare, energy, telecommunications, and urban planning. In engineering, they are used to optimize structural designs, energy systems, and control mechanisms. Machine learning applications highlight the integration of metaheuristics for hyperparameter tuning and feature selection, improving model performance. Supply chain management benefits from metaheuristics in solving routing and scheduling problems. Healthcare and bioinformatics leverage these algorithms for treatment planning and DNA sequencing. In smart cities, metaheuristics enhance urban planning and disaster management, while in energy systems, they optimize renewable energy generation and grid management. Across all fields, metaheuristics provide flexible, robust solutions for multi-objective and complex problems, making them essential in modern computational challenges.

\textbf{Optimization in Engineering:} Metaheuristics have been extensively used for optimizing complex engineering problems such as structural design, energy systems, and control systems. \citet{dhiman2021esa} introduced a hybrid bio-inspired optimization approach called the Emperor Penguin and Salp Swarm Algorithm (ESA), which mimics the huddling behavior of emperor penguins and the swarm dynamics of salps. The ESA's performance is evaluated on benchmark functions and engineering problems, demonstrating its ability to find optimal solutions compared to other metaheuristics. \citet{hayyolalam2020black} introduced the Black Widow Optimization Algorithm (BWO), a novel metaheuristic inspired by the mating behavior of black widow spiders, including a unique cannibalism stage for early convergence. The BWO is evaluated on 51 benchmark functions and three engineering design problems, demonstrating its effectiveness in solving complex, real-world optimization challenges with competitive results. \citet{bekdacs2019optimization} highlighted recent advances in the design optimization and applications of metaheuristic algorithms in civil engineering. It discusses the importance of optimization, reviews various metaheuristic techniques, and explores their effectiveness in solving complex, constrained design problems, with suggestions for further improvements.

\textbf{Machine Learning:} Metaheuristics are often employed to optimize hyperparameters in machine learning models, feature selection \cite{zebari2020comprehensive}, and in some cases for enhancing neural network training. In a recent study, \citet{akay2022comprehensive} provided an overview of deep neural network (DNN) architectures, optimization challenges, and how metaheuristic algorithms have been applied to automate tasks like architecture and hyper-parameter optimization. It categorizes encoding schemes, summarizes evolutionary operators, and discusses the pros, cons, and future directions of integrating metaheuristics with deep learning. \citet{talbi2021machine} explored the growing trend of integrating machine learning (ML) with metaheuristics to enhance their efficiency, effectiveness, and robustness, presenting a detailed taxonomy based on optimization components. He also highlights synergies between machine learning and metaheuristics, motivating researchers to further investigate this promising research direction while identifying open issues for future study. \citet{dokeroglu2022comprehensive} reviewed the most notable metaheuristic feature selection algorithms from the past two decades, highlighting their performance in exploration/exploitation, selection methods, transfer functions, fitness evaluations, and parameter settings. It also addresses current challenges and suggests future research directions for improving metaheuristic feature selection algorithms.

\textbf{Supply Chain and Logistics:} Metaheuristics like Tabu Search \cite{glover1998tabu} and Simulated Annealing \cite{van1987simulated} have been applied to solve complex problems related to routing, scheduling, and resource allocation in supply chains. \cite{song2020metaheuristics} addressed a vehicle routing problem (VRP) in cold chain logistics, incorporating dispatch time windows, multiple vehicle types, and varying energy consumption. An improved artificial fish swarm (IAFS) algorithm is developed, featuring a novel encoding approach for different vehicle types and improved preying, following, and customer satisfaction heuristics. \citet{rachih2019meta} reviewed and classified previous research on reverse logistics (RL), focusing on the use of metaheuristic approaches to solve complex optimization problems associated with reverse supply chain integration. The study highlights the efficiency and flexibility of metaheuristics in addressing RL challenges and explores future research opportunities for enhancing RL practices. \citet{govindan2019designing} addressed the growing need for sustainable supply chain models by integrating the triple bottom line (economic, environmental, and social impacts) into a distribution network model. The study solves a multi-product vehicle routing problem with time windows (MPVRPTW) using three hybrid swarm intelligence techniques—PSO, electromagnetism mechanism algorithm (EMA), and artificial bee colony (ABC)—each combined with variable neighbourhood search (VNS). 

\textbf{Healthcare and Bioinformatics:} In healthcare, metaheuristics have been used for optimizing medical imaging, treatment planning, and drug design. In bioinformatics, they are applied for DNA sequencing, protein structure prediction, and clustering of biological data. \citet{savanovic2023intrusion} addressed security challenges in IoT systems for Healthcare 4.0 by using machine learning algorithms optimized with a modified Firefly metaheuristic to detect issues. Experiments on synthetic IoT data demonstrated significant improvements, with SHapley Additive exPlanations (SHAP) analysis identifying key factors contributing to the problems, highlighting the potential of metaheuristics for sustainable healthcare solutions. \citet{nematzadeh2022tuning} presented a method for tuning hyperparameters of machine learning algorithms using metaheuristics. Testing 11 algorithms across diverse datasets, the results demonstrate that GWO outperforms other methods, significantly improving training performance and convergence compared to Exhaustive Grid Search (EGS), making it suitable for datasets with unknown distributions and complex algorithmic behavior. \citet{fathollahi2020set} addressed the challenges of home healthcare (HHC) operations by proposing a new mathematical formulation that incorporates innovative assumptions in the field. It introduces three new heuristics and a hybrid constructive metaheuristic to optimize nurse scheduling and routing. The algorithms' performance is validated against a developed lower bound using Lagrangian relaxation and is further analyzed through various criteria and sensitivity analyses to ensure the efficiency of the proposed model.

\textbf{Energy Systems:} The optimization of renewable energy sources, energy storage systems, and power grid management has been a growing application area for metaheuristics, particularly in solving multi-objective optimization problems. \citet{guven2022performance} investigated the techno-economics of a hybrid off-grid energy system integrating wind, solar, biomass gasifier, and fuel cell technologies, optimizing energy generation and storage through hydrogen. Using a Hybrid Firefly Genetic Algorithm, the system achieves optimal component sizing, minimizes annual costs, and demonstrates superior performance in terms of accuracy and calculation time compared to other algorithms. \citet{minai2021metaheuristics} focused on the role of metaheuristic optimization techniques in enhancing the efficiency and cost-effectiveness of power generation from renewable energy sources. They discussed the application of various approaches, such as PSO, Genetic Algorithms, and others, for optimizing systems like solar PV, battery storage, and wind farm design, aiming to improve productivity and reliability while minimizing costs. \citet{ikeda2015metaheuristic} explored the optimization of energy systems with battery and thermal energy storage using metaheuristic techniques like genetic algorithms, PSO, and cuckoo search. The results demonstrate that the proposed mutation-PSO (m-PSO) is the fastest method, while cuckoo search is the most accurate, offering significant computational advantages over traditional dynamic programming with minimal tolerance differences.

\textbf{Telecommunications and Networking:} Metaheuristics have been utilized to enhance the efficiency of network designs, improve wireless communication, and manage bandwidth allocation. \citet{iwendi2021metaheuristic} focused on optimizing energy consumption in IoT networks to extend network lifetime by selecting the most appropriate Cluster Head. Using a hybrid metaheuristic approach combining the Whale Optimization Algorithm (WOA) \cite{mirjalili2016whale} with Simulated Annealing (SA), the method improves performance based on metrics like residual energy and cost, demonstrating superiority over existing algorithms such as Artificial Bee Colony \cite{karaboga2009comparative}, Genetic Algorithm, and WOA. \citet{alizadeh2023optimized} proposed a novel hybrid method for short-term telecommunication traffic forecasting that combines statistical and machine learning approaches to model linear and nonlinear data components. Using a VARMA-LSTM-MLP forecaster and a hybrid metaheuristic algorithm of firefly and BAT for hyper-parameter optimization, the method demonstrates superior performance compared to existing approaches when tested on a real-world dataset from Tehran, Iran, in terms of mean squared error and mean absolute error. \cite{kostic2020social} explored the use of social network analytics and graph theory to analyze a large telecommunications network and identify key nodes that influence customer churn. By clustering nodes based on metrics such as in/out-degree and influence, the study demonstrates that the departure of specific nodes increases the likelihood of churn among their connected customers, allowing proactive churn prediction using top decile lift metrics. The method is versatile and can be applied in other fields where social connections drive churn.

\textbf{Finance, Economics, and Manufacturing:} Metaheuristics are applied in portfolio optimization, risk management, algorithmic trading, predicting financial market trends, job scheduling, inventory management, and quality control to optimize processes, reduce costs, and enhance productivity. Computational finance is an emerging field for metaheuristic algorithms, which are increasingly used to solve complex decision-making problems like portfolio optimization and risk management. \citet{doering2019metaheuristics} systematically reviewed the literature on these applications, identified links between portfolio optimization and risk management, and highlights future research trends.  Developing hybrid renewable energy systems is challenging due to the intermittency of renewables and complex design considerations. \citet{das2019techno} optimized the design of an off-grid hybrid system using metaheuristic algorithms, showing that the water cycle algorithm provides a slightly better solution than moth-flame optimization and Genetic Algorithm, resulting in a techno-economic design with a total net present cost of 0.813 million dollars. Reviewed solutions for modeling additive manufacturing process planning lack the necessary flexibility for hybrid additive/subtractive operations. \cite{rossi2020integration} proposed a system that extracts features from CAD, introduces operation and sequencing flexibility, and generates a precedence graph, which is optimized using Ant Colony Optimization to handle complex process planning and job shop scheduling effectively.

\textbf{Environmental and Ecological Modeling:} Metaheuristic techniques have been applied to model and simulate environmental systems such as climate change forecasting, water resource management, and species distribution optimization.  Recycled aggregate concrete (RAC) is gaining attention for its sustainability benefits in construction. \cite{liu2023mixture} developed a framework combining machine learning and metaheuristics to optimize RAC mixture proportions, with extreme gradient boosting providing the best compressive strength predictions. Proposed competitive mechanism-based multi-objective PSO algorithm effectively optimizes mechanical, economic, and environmental objectives, offering Pareto optimal solutions across multiple design scenarios. \cite{oliveira2020review} reviewed evolutionary and bio-inspired methods, such as simulated annealing, genetic algorithm, differential evolution, and PSO, and their applications to both single and multi-objective greenhouse control problems, highlighting current trends in this field.  Stochastic optimization methods, such as genetic algorithms, PSO, and tabu search (TS), are widely used for solving high-dimensional, nonlinear problems. \cite{roque2017metaheuristic} focused on the TS algorithm, highlighting its memory and adaptive features, and details its successful application to the dynamic optimization of a copolymerization reactor and inverse modeling of a biofilm reactor, demonstrating its efficacy in chemical and environmental processes.

\textbf{Smart Cities and Urban Planning:} Algorithms like Cuckoo Search \cite{yang2014cuckoo} and Artificial Bee Colony have been utilized in optimizing urban infrastructure planning, traffic management, waste management, and smart energy grids in the development of smart cities. Achieving sustainability in smart cities requires ongoing monitoring and adaptable systems. \cite{fanian2023cfmcrs} focused on wireless rechargeable sensor networks (WRSNs) for continuous monitoring and proposed a calibration fuzzy-metaheuristic clustering routing scheme that enhances energy efficiency, role management, and scheduling in WRSNs, outperforming existing methods in simulations by improving energy distribution, latency, and network lifespan, and these results were validated through ANOVA and post-hoc analysis. Evacuation planning is a critical multi-objective optimization problem in disaster management, often too complex for traditional methods. \cite{niyomubyeyi2020comparative} compared the performance of four classical multi-objective metaheuristic algorithms (AMOSA, MOABC, NSGA-II, and MSPSO) on an urban evacuation problem in Rwanda. Results show that AMOSA and MOABC provide high-quality solutions, while NSGA-II \cite{deb2002fast} is the fastest in terms of execution time and convergence speed. AMOSA, MOABC, and MSPSO demonstrated better repeatability, with potential improvements in MOABC suggesting its suitability for evacuation planning.

\textbf{Transportation Systems:} Optimization of transportation networks, including air traffic management, public transportation scheduling, and autonomous vehicle routing, has seen the application of metaheuristics like Harmony Search and Differential Evolution. \citet{sadeghi2019new} focused on solving the Fixed Charge Transportation Problem (FCTP) using fuzzy models for both fixed and variable costs, and introduces a new approach with the Whale Optimization Algorithm (WOA) alongside three other metaheuristics. Innovative representation techniques, such as spanning tree-based Prüfer number and priority-based representation, are employed, while the Taguchi method ensures the optimal performance and parameter tuning of the algorithms. \citet{juntama2022hyperheuristic} addressed the airspace capacity issue by minimizing traffic complexity using optimization techniques based on linear dynamical systems and traffic structuring methods such as departure time adjustment, trajectory deviation, and flight-level allocation. The proposed hyperheuristic framework, leveraging reinforcement learning, reduces air traffic complexity by 92.8\% in the French airspace and outperforms both random search and simulated annealing, with additional analysis considering time uncertainties for future capacity management. \citet{jamal2020intelligent} focuses on improving traffic signal control at isolated intersections, using metaheuristic methods like Genetic Algorithm (GA) and Differential Evolution (DE) to optimize signal timing and reduce vehicle delays. The results showed a 15-35\% reduction in travel time delays, with DE converging faster but GA delivering higher quality solutions. The performance of both algorithms was validated against the TRANSYT 7F tool, demonstrating the robustness of the proposed methods.

\textbf{Robotics and Drones:} Metaheuristics are being used in path planning, swarm robotics \cite{sacramento2019adaptive}, and robotic motion control, enabling robots to find optimal paths, avoid obstacles, and make decisions autonomously in dynamic environments. \citet{fong2015review} reviewed recent advances in metaheuristic algorithms applied to robotics, highlighting their impact on enhancing task performance, reliability, and cost-efficiency in collaborative robotic systems. It provides a taxonomy to guide robotics designers in leveraging these algorithms for improved coordination and interaction among reconfigurable, communicating robots. \citet{kiani2022adaptive} proposed two metaheuristic algorithms, Incremental Gray Wolf Optimization (I-GWO) and Expanded Gray Wolf Optimization (Ex-GWO), to solve the NP-hard problem of 3D path planning for autonomous robots in agriculture. The simulations demonstrate that the Ex-GWO algorithm achieves a 55.56\% better success rate in optimal path cost compared to other methods, effectively enabling robots to navigate collision-free paths and perform tasks like crop tracking efficiently. \citet{ab2020comparative} evaluates various metaheuristic algorithms for robot motion planning and compares their performance against traditional techniques like Dijkstra’s Algorithm and Rapidly Random Tree \cite{lavalle2001rapidly}. The results indicate that metaheuristic approaches are competitive with conventional methods, with Constricted PSO outperforming other metaheuristics in unknown environments.

\textbf{Cybersecurity:} Metaheuristic methods have been used to enhance intrusion detection systems \cite{ghanbarzadeh2023novel}, optimize firewall configurations, and design secure cryptographic protocols. \citet{salas2021metaheuristic}  investigated the use of metaheuristics to optimize artificial intelligence techniques for threat detection and attack optimization, analyzing 41 key articles from a comprehensive literature review. It finds that a significant focus is on reducing features during the training stage to improve real-time detection efficiency, with metaheuristics playing a crucial role in this process. \citet{diaba2023cyber} addressed vulnerabilities in power system communication protocols by proposing a metaheuristic-optimized Restricted Boltzmann Machine-based algorithm to enhance cyber-attack detection using deep learning techniques. Simulations demonstrate that this metaheuristic approach significantly outperforms traditional methods, achieving high accuracy in binary, three-class, and multi-class classification tasks. \citet{albakri2023metaheuristics}  introduced the Rock Hyrax Swarm Optimization with deep learning-based Android malware detection (RHSODL-AMD) model, which utilizes metaheuristic techniques for effective feature selection and API call analysis. Experimental results on the Andro-AutoPsy dataset demonstrate that the RHSODL-AMD model achieves a high accuracy of 99.05\% in distinguishing between benign and malicious applications.

\textbf{Software Engineering:} Metaheuristics have been used for optimizing software testing, refactoring code, and solving problems related to bug detection and software reliability. \citet{khan2021metaheuristic} contributed to software effort estimation by exploring metaheuristic algorithms for building a logical and acceptable parametric model. It introduces a Deep Neural Network (DNN) \cite{sze2017efficient} model optimized using Grey Wolf Optimizer (GWO) \cite{mirjalili2014grey} and StrawBerry (SB) algorithms \cite{merrikh2014numerical}, highlighting their effectiveness in estimation. Experimental results show GWO's superiority in accuracy and the improved performance of the proposed DNN model compared to previous approaches. \citet{zhu2021software} addressed software defect prediction by proposing an enhanced metaheuristic feature selection algorithm, using whale optimization \cite{mirjalili2016whale} and simulated annealing \cite{van1987simulated} to select fewer but relevant features. It also introduces a hybrid deep neural network, WSHCKE, combining CNN and kernel extreme learning machine, which boosts prediction performance, with experiments showing the superiority of both methods across 20 software projects. \citet{rhmann2022software} presented a software effort estimation model using a weighted ensemble of hybrid search-based metaheuristic algorithms, including firefly, black hole optimization, and genetic algorithms. Experiments demonstrate that this metaheuristic-based approach outperforms traditional machine learning models and their ensembles in predicting software development efforts.

\textbf{Water Resource Management:} Metaheuristic techniques have been employed in optimizing the allocation and management of water resources, ensuring sustainable usage in agriculture, industry, and urban areas.  \citet{maier2014evolutionary} reviewed the use of evolutionary algorithms and metaheuristics in optimizing water resource systems, highlighting the need for a more integrated approach to address common challenges across various applications. It calls for advances in fitness landscape understanding, problem formulation, and computational efficiency to enhance metaheuristics applications and support decision-making in complex, uncertain contexts. \citet{kumar2022state} provided a comprehensive review of heuristic and metaheuristic optimization techniques in water resource management, highlighting their effectiveness in addressing complex, non-linear, and multi-objective challenges. It emphasizes the benefits of hybrid and modified algorithms, offering valuable insights for researchers and practitioners in selecting optimal solutions for water resource problems. \citet{bhavya2023ant} reviewed the application of ant colony optimization algorithms in hydrology and hydrogeology, highlighting their effectiveness in managing complex water resource problems. Despite their potential and improvements through hybrid techniques, challenges such as incorporating uncertainty and resolving issues related to dimensionality, convergence, and stability remain areas for future research.

\section{Discussions on Challenges and Open Research Issues}\label{section5}

In this section, we tried to touch upon some important topics under the titles Similarity analysis of our selected algorithms with other recent/classical metaheuristics, parameter sensitivity and tuning, local optima, and binary encoding.

\subsection{Similarity analysis of our selected algorithms with other recent/classical metaheuristics}

Selected metaheuristic algorithms are criticized because of their similarities with other metaheuristics. Although the mathematical formulations they propose are different, their solution strategies are similar to some metaheuristics developed before. Table \ref{similarities} presents the names of similar metaheuristics in terms of hunting, interaction, exploration, food-finding, and nest-switching behaviours. PSO, Grey-Wolf, and Whale optimization are the most simulated approaches.

\begin{table}[h!]
	\caption{Comparison of similarities between metaheuristic algorithms that are introduced between 2019 and 2024.}	\label{similarities}
	\begin{center}
	\scalebox{0.74}{
		\begin{tabular}{lll}
\hline
\textbf{New Metaheuristic} & \textbf{Similar Metaheuristic} & \textbf{Reason for Similarity} \\
\hline 
Harris hawk optimization & Grey Wolf  &   Social interaction and hunting strategies  \\
Butterfly optimization & Bacterial Foraging & Natural food-finding behaviors. \\ 
Gradient-based optimizer  &   Differential Evolution &  Improves the solution progressively.  \\
Slime Mould algorithm & Harmony Search & Exhibits sensory interactions and organizational behaviors. \\  
Marine Predators Algorithm & Grey Wolf & Social hunting behaviors. \\ 
Equilibrium optimizer & PSO & Employs global search and social interaction mechanisms. \\  
Aquila Optimizer & Harris Hawk & Shares similar hunting and social behaviors. \\  
Seagull Optimization & PSO & Employs similar social interaction and exploration mechanisms. \\  
Manta ray foraging optimization & Whale Optimization & Similar hunting strategies. \\  
Chimp optimization & Cuckoo Search & Similar nest-switching and social interaction strategies. \\  
Squirrel Search Algorithm  & PSO &   Using individual and collective strategies. \\
Henry gas solubility optimization  &  Cuckoo Search Algorithm & Using a combination of exploration and exploitation  \\
Archimedes optimization algorithm & Genetic Algorithm & Adopts principles from genetic algorithms. \\  
Tunicate Swarm Algorithm & PSO & Based on social interaction mechanisms. \\  
Honey Badger Algorithm & Bat Algorithm & Shares sensory tracking and hunting behaviors. \\  
Mayfly optimization & Firefly Algorithm  & The swarm is attracted to the best solutions   \\
African vultures optimization & Grey Wolf & Utilizes social interaction and hunting strategies. \\  
Golden jackal optimization & Teaching-learning-Based & Based on learning and interaction processes. \\  
Dung beetle optimizer & Gravitational Search & Similar methods for carrying loads. \\  
Coati Optimization & Social Spider & Based on social interactions. \\  
Chaos Game Optimization  &  Artificial Bee Colony &  Chaotic dynamics to guide the population   \\
Beluga whale optimization & Whale Optimization & Involves similar interaction and exploration strategies. \\  
Gazelle optimization & Firefly Algorithm & Utilizes principles of attraction and social interactions. \\  
\hline
\end{tabular}
}
\end{center}
\end{table}
\vspace{-0.5cm}

\subsection{Parameter Sensitivity and Tuning}

Parameter setting is a critical aspect of metaheuristic algorithms, as it can significantly influence their performance and ease of use \citep{huang2019survey}. The selected algorithms—such as African Vultures Optimization, Aquila Optimizer, Archimedes Optimization, Beluga Whale Optimization, Butterfly Optimization, Chimp Optimization, Coati Optimization, Dung Beetle Optimization, Equilibrium Optimizer, Gazelle Optimization, Honey Badger Algorithm, Manta Ray Foraging Optimization, Marine Predators Algorithm, Prairie Dog Optimization, Slime Mould Algorithm, and Tunicate Swarm showcase various strategies in parameter handling.

Some metaheuristics aim to minimize or eliminate the need for parameter tuning, simplifying their use and making them more accessible to non-expert users. Algorithms like the Equilibrium Optimizer and Snake Optimizer are examples that require minimal parameter adjustments, focusing on simplicity and robustness. The Slime Mould Algorithm, while involving a few parameters, adapts its behavior dynamically, making it less dependent on meticulous tuning. The key advantage of parameterless or minimally parameterized algorithms is their ease of application across different problem domains without needing extensive parameter optimization. This feature reduces the overhead of trial-and-error testing, making them highly suitable for practical, time-sensitive applications. However, the drawback is that parameterless algorithms might not achieve peak performance in highly specialized or complex problem scenarios where fine-tuning could unlock greater optimization potential. These algorithms tend to be designed with general settings that may not fully leverage specific problem characteristics.

In contrast, algorithms like Aquila Optimizer, Butterfly Optimization, and Honey Badger Algorithm involve multiple parameters to control different aspects of their search behavior, such as exploration-exploitation balance and convergence rate. For instance, Aquila Optimizer has parameters to modulate different flight patterns, allowing it to adapt its strategy as needed. This level of control provides users the ability to fine-tune the algorithm to fit specific problem constraints, leading to potentially superior results in complex scenarios.

The advantage of such highly parameterized algorithms is their adaptability and potential for high performance when properly configured. By tweaking parameters, users can optimize these algorithms for different types of landscapes, increasing their versatility and effectiveness. However, this flexibility comes at a cost: the need for extensive experimentation or sophisticated tuning techniques (e.g., grid search or meta-optimization) to identify the best parameter settings. This requirement can be time-consuming and may limit the algorithm’s practicality, especially for those who lack experience or computational resources.

Algorithms like Marine Predators Algorithm and Manta Ray Foraging Optimization strike a balance between having some key parameters that allow for moderate customization without overwhelming the user with complexity. These algorithms often feature straightforward parameter interactions that simplify the tuning process. The RIME algorithm is another example, blending minimal parameter settings with adaptive behavior to improve convergence. Meanwhile, certain niche algorithms like Dung Beetle Optimization and Gazelle Optimization include parameters that emulate real-world animal behaviors. This bio-inspired aspect can provide intuitive insights into parameter adjustments, making them more approachable than more abstract methods.

Parameterless algorithms offer significant advantages in terms of ease of use and fast deployment. They are ideal for general problems where extensive customization is unnecessary \citep{dushatskiy2024parameterless}. However, they may be less effective in domains requiring specific performance optimizations. Highly parameterized algorithms, conversely, can excel in diverse and challenging problem spaces when tailored appropriately, but at the expense of higher computational and expertise costs. In conclusion, the choice between parameterless and parameter-rich metaheuristic algorithms depends on the problem's complexity, user expertise, and available resources. For practitioners who need fast, out-of-the-box solutions, parameterless methods are advantageous. For those seeking maximum performance and flexibility, investing in parameter tuning for more complex algorithms can yield better outcomes.

\subsection{Escaping from local optima}

Overcoming local optima is a crucial challenge in optimization algorithms, as it prevents convergence to the global optimum \citep{rego2007local,rajabi2023stagnation}. Several techniques have been developed to enhance the ability of algorithms to escape local optima, ensuring more efficient and robust solutions to complex problems. One approach is combining global and local search methods. The global search explores large regions of the solution space, while local search focuses on refining solutions. This hybrid approach increases the chances of escaping local optima by allowing for broader exploration and more focused refinement. Additionally, incorporating memory mechanisms that track previous solutions prevents revisiting the same local optima and helps guide the search toward better solutions.

Introducing randomization is another key strategy. By allowing the algorithm to accept worse solutions with some probability, it diversifies the search and reduces the likelihood of getting stuck in local optima. Modifying the cooling schedule in probabilistic methods can also improve exploration by adjusting how the algorithm transitions from exploration to exploitation, fostering better balance in the search process. Maintaining diversity within the population is essential to avoid local optima. Techniques like crowding and niche sharing help preserve variation, ensuring that the algorithm continues exploring different regions of the solution space. Adaptive methods that adjust population size or selection pressure further enhance diversity, promoting exploration without premature convergence. Finally, memory-based approaches store elite solutions to guide the search and avoid redundant exploration. Hybridizing optimization methods and incorporating learning techniques, such as reinforcement learning, can further improve the algorithm’s ability to escape local optima by dynamically adjusting the search process based on prior results.

\subsection{Binary encoding}

The process of converting continuous optimization variables into binary variables is commonly referred to as "discretization" or "binary encoding" \citep{liu2002discretization}. In the context of metaheuristic algorithms, this transformation allows continuous solutions to be effectively applied in binary decision-making scenarios, such as feature selection or other combinatorial optimization problems.

By employing methods like the S-shape or V-shape functions during this discretization process, algorithms can maintain a balance between exploration of the solution space and adherence to binary constraints, thereby improving overall optimization performance \citep{sharafi2021opposition}. Techniques like S-shape and V-shape functions are commonly employed to convert continuous models into binary formats. This conversion is crucial when dealing with binary decision-making problems, such as feature selection, where the goal is to determine the inclusion or exclusion of specific features.

The S-shape function typically maps continuous values into a binary space using a sigmoid-like curve. This approach ensures a smooth transition, allowing for gradual changes in the decision-making process. The S-shape function is particularly useful when a soft transition between decisions is required, enabling a more refined exploration of the solution space.

On the other hand, the V-shape function provides a more abrupt transition from continuous to binary values. It effectively enforces a strict threshold, where values below a certain point are classified as one binary state (e.g., 0), and those above are classified as another (e.g., 1). This method is beneficial when clear-cut decisions are necessary, as it reduces ambiguity in the selection process.

Both techniques enhance the adaptability of metaheuristic algorithms by allowing them to effectively navigate continuous landscapes while meeting the binary constraints inherent in specific optimization problems. Ultimately, the choice between S-shape and V-shape functions depends on the specific requirements and characteristics of the problem at hand.

\section{Conclusion}\label{section6}

The development of new metaheuristic algorithms is likely to continue as these innovative studies have a high likelihood of publication and appeal to a broad range of applications. Metaheuristics require novel approaches to address the diversity of optimization problems, offering researchers opportunities to create unique algorithms. Furthermore, these algorithms are gaining popularity for providing more efficient solutions across various fields, making them widely applicable. Hybrid algorithms and integrations with machine learning, in particular, present attractive solutions for tackling increasingly complex problems, which further drives the number of studies and publications in this area. Selecting enduring and effective algorithms from hundreds of new metaheuristics will remain a significant challenge for researchers. The success of these algorithms will be evaluated based on their ability to adapt to a wide range of problems, demonstrate effective performance, and gain broad acceptance.

Our article is likely to make a strong impact by providing a comprehensive summary of the standout metaheuristic algorithms from the past six years. Highlighting key developments and trends in this evolving field will offer valuable insights for researchers and practitioners navigating the landscape of optimization techniques.

The future of metaheuristics is bright, driven by advances in machine learning and generative AI. Integrating ML can enhance metaheuristics with adaptive, self-tuning capabilities, making them more accessible and powerful across diverse applications. Generative AI can assist in exploring solution spaces creatively, improving initialization and diversity strategies. Hybrid metaheuristics combining multiple algorithms are poised to tackle increasingly complex, high-dimensional problems in engineering, healthcare, and logistics problems. By embracing parallel and cloud computing, metaheuristics will achieve faster, scalable solutions, establishing them as essential tools for optimization in AI-driven and data-intensive industries of the future.

Future work in parallel metaheuristics promises to address scalability, efficiency, and convergence speed for complex optimization problems. By distributing computation across multiple processors, parallel metaheuristics reduce execution time and enhance solution quality, especially in large-scale and real-time applications. Developing adaptive parallel frameworks that dynamically adjust parameters and optimize resource usage could further improve performance. Moreover, combining parallelism with hybrid metaheuristic approaches may yield robust solutions by leveraging the complementary strengths of different algorithms. Future studies can also explore parallel implementations on GPU clusters and cloud infrastructures to handle high-dimensional data and optimize energy and resource management.\\

\textbf{Acknowledgement}

We would like to thank Özlem Tekdemir Dökeroglu for her artistic illustrations in Figure \ref{Slime}.

\appendix
\section{Other recent metaheuristic algorithms proposed between 2019 and 2024}\label{appendix}

\begin{center}
\begin{longtable}{llr}
\caption{The metaheuristic algorithms developed between 2019 and 2024 (sorted by the name of the algorithms).}\label{metaheuristics-2}\\
\hline
\ \textbf{Metaheuristic} & \textbf{Year} & \textbf{\#citations} \\ 
\hline
 Sunflower Optimization \citep{gomes2019sunflower} 	&	2019	&	1580	\\
 Black Widow Optimization Algorithm \citep{hayyolalam2020black} 	&	2020	&	1330	\\ 
 Artificial ecosystem-based optimization \citep{zhao2020artificial} 	&	2020	&	1200	\\
 Political Optimizer \citep{askari2020political} 	&	2020	&	1020	\\
 Pelican Optimization Algorithm \citep{trojovsky2022pelican} 	&	2022	&	967	\\
 Artificial gorilla troops optimizer \citep{abdollahzadeh2021artificial} 	&	2021	&	967	\\
 Sailfish Optimizer \citep{shadravan2019sailfish} 	&	2019	&	953	\\
 Snake Optimizer \citep{hashim2022snake} 	&	2022	&	916	\\
 Remora optimization \citep{jia2021remora} 	&	2021	&	816	\\
 Artificial rabbits optimization \citep{wang2022artificial} 	&	2022	&	815	\\
 RUNge Kutta optimizer \citep{ahmadianfar2021run} 	&	2021	&	801	\\
 Chameleon Swarm Algorithm \citep{braik2021chameleon} 	&	2021	&	790	\\
 Flow Direction Algorithm \citep{karami2021flow} 	&	2021	&	765	\\
 Wild Horse Optimizer  \citep{naruei2022wild} 	&	2022	&	745	\\
 Barnacles Mating Optimizer \citep{sulaiman2020barnacles} 	&	2020	&	720	\\
 Horse Herd \citep{miarnaeimi2021horse} 	&	2021	&	703	\\
 Bald eagle search optimiZation \citep{alsattar2020novel} 	&	2020	&	667	\\
 Deer Hunting Optimization Algorithm \citep{brammya2019deer} 	&	2019	&	605	\\
 Red fox optimization  \citep{polap2021red} 	&	2021	&	594	\\
 Gaining Sharing Knowledge Based Algorithm \citep{mohamed2020gaining} 	&	2020	&	593	\\
 Prairie dog optimization \citep{ezugwu2022prairie} 	&	2022	&	584	\\
 Transient Search Optimization \citep{qais2020transient} 	&	2020	&	570	\\
 Water strider algorithm \citep{kaveh2020water} 	&	2020	&	557	\\
 Dandelion Optimizer \citep{zhao2022dandelion} 	&	2022	&	554	\\
 White Shark Optimizer \citep{braik2022white} 	&	2022	&	547	\\
 Golden eagle optimizer \citep{mohammadi2021golden} 	&	2021	&	542	\\
 COOT bird \citep{naruei2021new} 	&	2021	&	510	\\
 Capuchin Search Algorithm  \citep{braik2021novel} 	&	2021	&	510	\\
 Ebola Optimization Search Algorithm \citep{oyelade2022ebola} 	&	2022	&	491	\\
 Tuna Swarm Optimization \citep{xie2021tuna} 	&	2021	&	486	\\
 Coronavirus Herd Immunity Optimizer  \citep{al2021coronavirus} 	&	2021	&	459	\\
 Jellyfish in ocean \citep{chou2021novel} 	&	2021	&	446	\\
 Atomic orbital search \citep{azizi2021atomic} 	&	2021	&	415	\\
 Spider wasp optimizer \citep{abdel2023spider} 	&	2023	&	397	\\
 RIME: A physics-based optimization \citep{su2023rime} 	&	2023	&	375	\\
 Mountain Gazelle Optimizer \citep{abdollahzadeh2022mountain} 	&	2022	&	362	\\
 Fire Hawk Optimizer \citep{azizi2023fire} 	&	2023	&	359	\\
 Student psychology based optimization \citep{das2020student} 	&	2020	&	359	\\
 Pathfinder algorithm \citep{yapici2019new} 	&	2019	&	329	\\
 Osprey optimization algorithm \citep{dehghani2023osprey} 	&	2023	&	304	\\
 War Strategy Optimization Algorithm \citep{ayyarao2022war} 	&	2022	&	303	\\
 Lichtenberg algorithm \citep{pereira2021lichtenberg} 	&	2021	&	282	\\
 Poor and rich optimization algorithm \citep{moosavi2019poor} 	&	2019	&	270	\\
 Emperor Penguins Colony \citep{harifi2019emperor} 	&	2019	&	267	\\
 Shuffled Shepherd Optimization \citep{kaveh2020shuffled} 	&	2020	&	264	\\
 Rain optimization algorithm \citep{moazzeni2020rain} 	&	2020	&	253	\\
 Kepler optimization algorithm \citep{abdel2023kepler} 	&	2023	&	244	\\
 Adolescent Identity Search Algorithm \citep{bogar2020adolescent} 	&	2020	&	244	\\
 Forensic Based Investigation \citep{chou2020fbi} 	&	2020	&	234	\\
 Dingo Optimizer \citep{bairwa2021dingo} 	&	2021	&	232	\\
 Starling murmuration optimizer \citep{zamani2022starling} 	&	2022	&	228	\\
 Gannet optimization algorithm \citep{pan2022gannet} 	&	2022	&	223	\\
 Sea-horse optimizer \citep{zhao2023sea} 	&	2023	&	222	\\
 Growth Optimizer \citep{zhang2023growth} 	&	2023	&	219	\\
 Levy flight distribution \citep{houssein2020levy} 	&	2020	&	211	\\
 Giza Pyramids Construction \citep{harifi2021giza} 	&	2021	&	206	\\
 Giza Pyramids Construction \citep{harifi2021giza} 	&	2021	&	205	\\
 Tasmanian Devil Optimization \citep{dehghani2022tasmanian} 	&	2022	&	201	\\
 Orca Predation Algorithm \citep{jiang2022orca} 	&	2022	&	194	\\
 Binary Chimp Optimization Algorithm \citep{wang2021binary} 	&	2021	&	193	\\
 Nomadic People Optimizer \citep{salih2020new} 	&	2020	&	191	\\
 Cheetah optimizer \citep{akbari2022cheetah} 	&	2022	&	189	\\
 Golden ratio optimization \citep{nematollahi2020novel} 	&	2020	&	184	\\
 Material Generation Algorithm \citep{talatahari2021material} 	&	2021	&	174	\\
 Crystal Structure Algorithm (CryStAl) \citep{talatahari2021crystal} 	&	2021	&	167	\\
 Stochastic Paint Optimizer \citep{kaveh2022stochastic} 	&	2022	&	166	\\
 Waterwheel Plant Algorithm \citep{abdelhamid2023waterwheel} 	&	2023	&	161	\\
 Dynamic differential annealed optimization \citep{ghafil2020dynamic} 	&	2020	&	158	\\
 Nutcracker optimizer \citep{abdel2023nutcracker} 	&	2023	&	157	\\
 Carnivorous Plant  Algorithm \citep{ong2021carnivorous} 	&	2021	&	153	\\
 Strawberry algorithm \citep{minh2023termite} 	&	2021	&	153	\\
 Liver Cancer Algorithm \citep{houssein2023liver} 	&	2023	&	148	\\
 Energy valley optimizer \citep{azizi2023energy} 	&	2023	&	146	\\
 Subtraction-Average-Based Optimizer \citep{trojovsky2023subtraction} 	&	2023	&	143	\\
 Tiki-taka algorithm \citep{ab2021tiki} 	&	2021	&	143	\\
 Newton Metaheuristic Algorithm \citep{gholizadeh2020new} 	&	2020	&	143	\\
 Walrus Optimization Algorithm \citep{trojovsky2023new} 	&	2023	&	138	\\
 Mother optimization algorithm \citep{matouvsova2023mother} 	&	2023	&	135	\\
 Youngs double-slit experiment optimizer \citep{merrikh2014numerical} 	&	2023	&	129	\\
 Snow ablation optimizer \citep{deng2023snow} 	&	2023	&	128	\\
 Interactive autodidactic school \citep{jahangiri2020interactive} 	&	2020	&	122	\\
 Chaotic vortex search algorithm \citep{gharehchopogh2022chaotic} 	&	2022	&	119	\\
 Light Spectrum Optimizer \citep{abdel2022light} 	&	2022	&	114	\\
 Genghis Khan shark optimizer \citep{hu2023genghis} 	&	2023	&	111	\\
 Komodo Mlipir Algorithm 	&	2022	&	111	\\
 Immune Plasma Algorithm \citep{aslan2020immune} 	&	2020	&	110	\\
 Giant Trevally Optimizer \citep{sadeeq2022giant} 	&	2022	&	108	\\
 Termite life cycle optimizer \citep{minh2023termite} 	&	2023	&	106	\\
 Mayfly in Harmony \citep{bhattacharyya2020mayfly} 	&	2020	&	105	\\
 Color Harmony Algorithm \citep{zaeimi2020color} 	&	2020	&	97	\\
 Alpine skiing optimization \citep{yuan2022alpine} 	&	2022	&	96	\\
 Sinh cosh optimizer \citep{trojovsky2022siberian} 	&	2023	&	84	\\
 Special Relativity Search \citep{goodarzimehr2022special} 	&	2022	&	84	\\
 Crested Porcupine Optimizer \citep{abdel2024crested} 	&	2024	&	81	\\
 Aphid-Ant Mutualism \citep{eslami2022aphid} 	&	2022	&	78	\\
 Caledonian crow learning algorithm \citep{al2020new} 	&	2020	&	78	\\
 Chaotic marine predators algorithm \citep{garip2024chaotic} 	&	2024	&	75	\\
 Chernobyl disaster optimizer \citep{shehadeh2023chernobyl} 	&	2023	&	75	\\
 SHADE WOA \citep{chakraborty2021shade} 	&	2021	&	72	\\
 Peafowl optimization \citep{wang2022novel} 	&	2022	&	70	\\
 Great Wall Construction \citep{guan2023great} 	&	2023	&	64	\\
 Mountaineering Team-Based Optimization \citep{faridmehr2023mountaineering} 	&	2023	&	61	\\
 Firebug Swarm Optimization \citep{noel2021new} 	&	2021	&	61	\\
 Elephant clan optimization \citep{jafari2021elephant} 	&	2021	&	56	\\
 Ludo game optimizer \citep{singh2019ludo} 	&	2019	&	56	\\
 Meerkat optimization algorithm \citep{xian2023meerkat} 	&	2023	&	55	\\
 Siberian tiger optimization \citep{bai2023sinh} 	&	2023	&	55	\\
 Bear smell search algorithm \citep{ghasemi2020novel} 	&	2020	&	54	\\
 Human urbanization algorithm \citep{ghasemian2020human} 	&	2020	&	53	\\
 Golf optimization algorithm \citep{montazeri2023golf} 	&	2023	&	52	\\
 Artificial Feeding Birds \citep{lamy2019artificial} 	&	2019	&	52	\\
 Solar System Algorithm \citep{zitouni2020solar} 	&	2020	&	50	\\
 Lemurs Optimizer \citep{abasi2022lemurs} 	&	2022	&	49	\\
 Fick's Law Algorithm \citep{hashim2023fick} 	&	2023	&	47	\\
 Trees Social Relations Optimization \citep{alimoradi2022trees} 	&	2022	&	45	\\
 Artificial lizard search optimization \citep{kumar2021artificial} 	&	2021	&	42	\\
 Owl Optimization Algorithm \citep{de2019metaheuristic} 	&	2019	&	38	\\
 Attack-Leave Optimizer \citep{kusuma2023attack} 	&	2023	&	33	\\
  Billiards Optimization Algorithm \citep{givi2023billiards} 	&	2023	&	33	\\
 Squid game optimizer \citep{azizi2023squid} 	&	2023	&	30	\\
 Running city game optimizer \citep{ma2023running} 	&	2023	&	28	\\
 Golden-Sine dynamic marine predator algorithm \citep{han2022golden} 	&	2022	&	26	\\
 Blue monkey \citep{mahmood2019blue} 	&	2019	&	24	\\
 Innovative gunner \citep{pijarski2019new} 	&	2019	&	22	\\
 Hiking Optimization Algorithm \citep{oladejo2024hiking} 	&	2024	&	20	\\
One-to-One-Based Optimizer \citep{dehghani2023oobo} 	&	2023	&	20	\\
 Artificial Protozoa Optimizer \citep{wang2024artificial} 	&	2024	&	18	\\
 Al-Biruni Earth Radius \citep{el2023biruni} 	&	2023	&	17	\\
 Ameliorated Young's double-slit experiment optimizer \citep{hu2023iydse} 	&	2023	&	16	\\
 Geometric Octal Zones Distance Estimation (GOZDE) \citep{kuyu2022gozde} 	&	2022	&	15	\\
 Flood algorithm \citep{ghasemi2024flood} 	&	2024	&	12	\\
 Puma optimizer \citep{abdollahzadeh2024puma} 	&	2024	&	7	\\
 Blood-sucking leech optimizer \citep{bai2024blood} 	&	2024	&	5	\\
 Piranha predation optimization algorithm \citep{zhang2024piranha} 	&	2024	&	1	\\
 Polar Fox \citep{ghiaskar2024polar} 	&	2024	&	1	\\
\hline
\end{longtable}
\end{center}


\bibliography{thebibliography}

\end{document}